\documentclass[10pt,twocolumn,letterpaper]{article}

\usepackage{cvpr}              
\usepackage{algorithm}
\usepackage{algorithmic}
\usepackage{multirow}
\usepackage{makecell}
\usepackage{booktabs}

\usepackage[table]{xcolor} 
\usepackage{mathtools}
\usepackage{graphicx}
\usepackage{nicefrac}
\usepackage[accsupp]{axessibility}  
\definecolor{cvprblue}{rgb}{0.21,0.49,0.74}
\definecolor{lightblue}{rgb}{0.9, 0.95, 1} 

\usepackage[pagebackref,breaklinks,colorlinks,allcolors=cvprblue]{hyperref}
\usepackage{pifont} 
\newcommand{\cmark}{\ding{51}}%
\newcommand{\xmark}{\ding{55}}%
\newcommand{\fmark}{\ding{109}}

\def\paperID{1298} 
\def\confName{CVPR}
\def\confYear{2026}

\definecolor{myorange}{HTML}{EE8980} 
\definecolor{mypink}{RGB}{255, 150, 200}
\definecolor{myblue}{RGB}{80, 180, 255}
\definecolor{mygray}{HTML}{9CB898}

\def\modelname{Harmony}

\newcommand{\yr}{\textcolor{black}}
\newcommand{\hut}{\textcolor{black}}
\newcommand{\hutnew}{\textcolor{black}}

\title{Harmony: Harmonizing Audio and Video Generation \\ through Cross-Task Synergy}

\author{
Teng Hu$^1$\footnotemark[1]\thanks{Equal Contribution}
\quad Zhentao Yu$^2$\footnotemark[1]
\quad Guozhen Zhang$^2$
\quad Zihan Su$^1$
\quad Zhengguang Zhou$^2$\\
\quad Youliang Zhang$^2$
\quad Yuan Zhou$^2$
\quad Qinglin Lu$^2$ 
\quad Ran Yi$^1$\thanks{Corresponding author.} \\
\normalsize $^1$Shanghai Jiao Tong University \quad $^2$Tencent Hunyuan\\
{\tt\small Project page: \href{https://sjtuplayer.github.io/projects/Harmony}{\textcolor{magenta}{https://sjtuplayer.github.io/projects/Harmony}}}
\\
}

\begin{document}

\twocolumn[{%
\maketitle
\begin{figure}[H]
\vspace{-0.3in}
\hsize=\textwidth 
\centering
\includegraphics[width=0.95\textwidth]{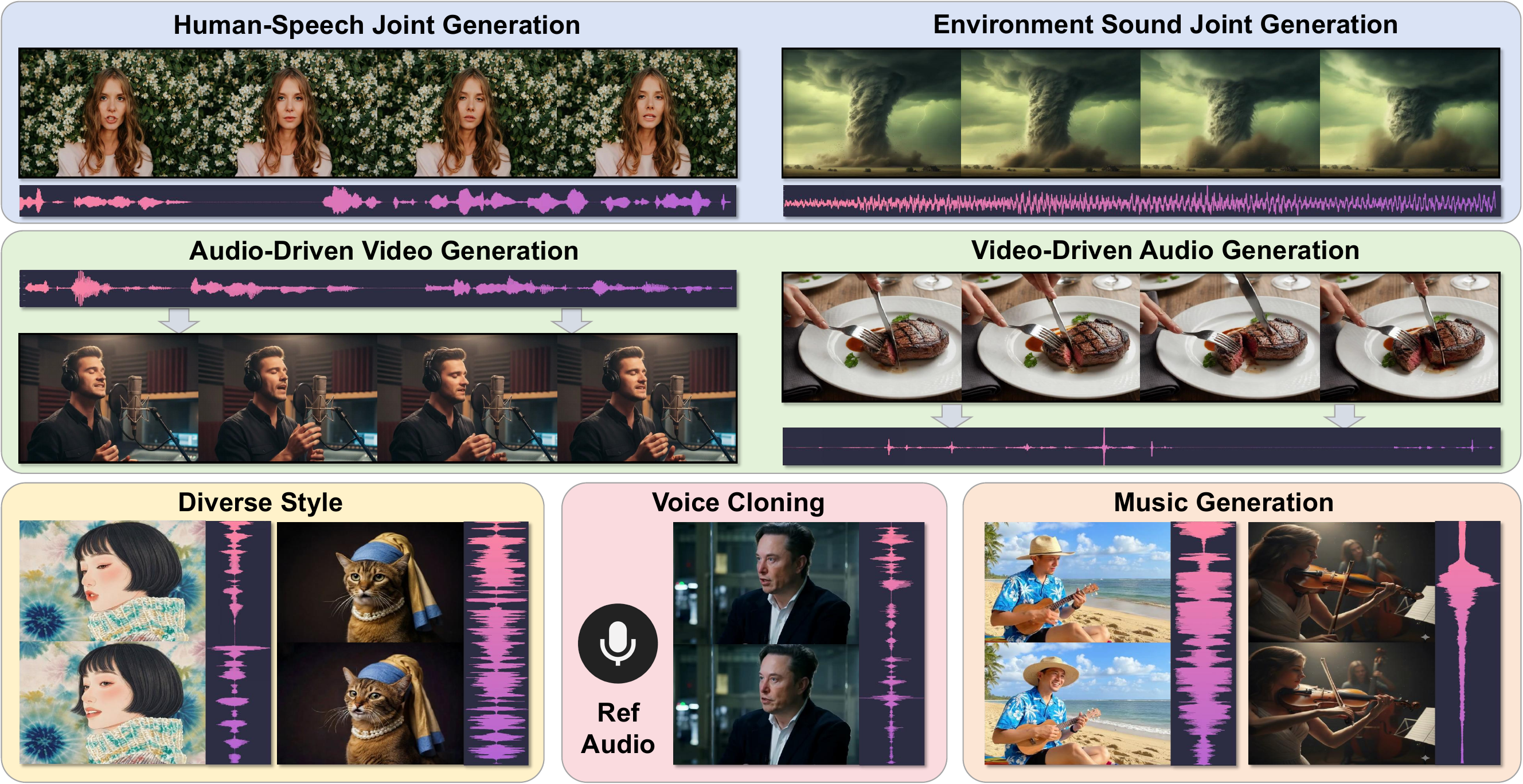}
\vspace{-0.05in}
\caption{\textbf{Harmony} employs a cross-task synergy training strategy to achieve robust audio-visual synchronization. This versatile framework supports multiple generation paradigms, including joint audio-video synthesis as well as audio-driven and video-driven generation, while also demonstrating strong generalizability to diverse audio types (e.g., music) and visual styles.
}
\vspace{-0.05in}
\end{figure}
}]

\maketitle

{%
\renewcommand{\thefootnote}{}%
\footnotetext{\textsuperscript{*} Equal Contribution.}%
\footnotetext{\textsuperscript{\dag} Corresponding author.}%
}

\begin{abstract}

The synthesis of synchronized audio-visual content is a key challenge in generative AI, with open-source models facing challenges in robust audio-video alignment. 
Our analysis reveals that this issue is rooted in three fundamental challenges of the joint diffusion process: (1) \textbf{Correspondence Drift}, where concurrently evolving noisy latents impede stable learning of alignment; (2) inefficient global attention mechanisms that fail to capture fine-grained temporal cues; and (3) the intra-modal bias of conventional Classifier-Free Guidance (CFG), which enhances conditionality but not cross-modal synchronization.
To overcome these challenges, we introduce \textbf{Harmony}, a novel framework that mechanistically enforces audio-visual synchronization. We first propose a \textbf{Cross-Task Synergy} training paradigm to mitigate drift by leveraging strong supervisory signals from audio-driven video and video-driven audio generation tasks. Then, we design a \textbf{Global-Local Decoupled Interaction Module} for efficient and precise temporal-style alignment. Finally, we present a novel \textbf{Synchronization-Enhanced CFG (SyncCFG)} that explicitly isolates and amplifies the alignment signal during inference. Extensive experiments demonstrate that Harmony establishes a new state-of-the-art, significantly outperforming existing methods in both generation fidelity and, critically, in achieving fine-grained audio-visual synchronization.

\end{abstract}


\section{Introduction}
\label{sec:intro}

The unified synthesis of audio and video represents a pivotal frontier in contemporary generative AI, with profound implications for content creation, digital avatars, and immersive virtual worlds. Industry-leading proprietary models, such as Veo 3~\cite{veo3_website} and Sora 2~\cite{sora2_website}, have set a high benchmark, delivering outputs with remarkable fidelity and demonstrating substantial practical utility. However, a significant gap persists between these closed-source systems and the capabilities of existing open-source methods~\cite{shan2025hunyuanfoley,cheng2025mmaudio,peng2025omnisync}. A fundamental challenge, in particular, remains largely unsolved in the open-source community: achieving precise and harmonious audio-visual alignment.

While recent open-source models have made strides in generation quality, they often struggle with robust audio-visual synchronization. Recent explorations into end-to-end \textbf{joint audio-video generation}~\cite{liu2025javisdit,ruan2023mmdiffusion,ishii2024simple,haji2024av,low2025ovi,wang2025universe} underscore this limitation.
\hut{Specifically, these methods often exhibit specialized limitations: many are confined to generating ambient sounds and fail to synthesize natural human speech~\cite{liu2025javisdit,ruan2023mmdiffusion,ishii2024simple,haji2024av}; while others, such as JAM-Flow~\cite{kwon2025jam}, focus solely on speech generation but lack the capability to generate environmental sounds.}
Even among more general models, 
\hutnew{Ovi~\cite{low2025ovi} exhibits deficiencies in \textit{robust alignment}, while UniVerse-1~\cite{wang2025universe} suffers from poor \textit{audio-video synchronization}.}
\yr{Table~\ref{tab:related_work_comparison} presents a capability comparison.}
These shortcomings reveal a critical gap in current research: few methods investigate the root causes of \yr{audio-video} misalignment from a methodological standpoint. 
Consequently, there remains a lack of highly generalizable and well-aligned audio-video \yr{joint} generation methods.
\hutnew{\textit{This leaves a significant void in the open-source landscape for a unified framework capable of generating a comprehensive audio spectrum—from ambient sounds to human speech—while maintaining precise audio-visual harmony.}}

In this work, we posit that the difficulty in achieving robust synchronization stems from three fundamental challenges inherent to the joint diffusion process. 
\textit{\yr{(1)}} During joint generation, both modalities are progressively denoised from pure noise. In the early, highly stochastic stages, attempting to align two concurrently evolving, \yr{highly noisy} latents \yr{causes} a phenomenon we term \textbf{Correspondence Drift}, where the optimal mapping continuously shifts, impeding stable learning. 
\textit{\yr{(2)}} 
\hut{Audio-visual synchronization presents a fundamental architectural tension between two competing objectives: precise, frame-level temporal alignment (\textit{e.g.}, lip movements) and holistic, global style consistency (\textit{e.g.}, emotional tone). Existing designs, often relying on a single, monolithic mechanism like global cross-attention, conflate these distinct goals, forcing the model into a suboptimal trade-off where neither objective is fully achieved.}
\textit{\yr{(3)}} Conventional Classifier-Free Guidance (CFG)~\cite{ho2022classifier} operates by amplifying conditioning signals for each modality in isolation. Consequently, it does not inherently promote or enhance the crucial cross-modal correspondence between the generated audio and video.

To overcome these challenges, we propose \textbf{Harmony}, a novel \yr{joint audio-video generation} framework designed to generate highly synchronized audio-\yr{video} content 
\yr{with cross-task synergy.}
Harmony's design is centered on three core innovations, each targeting one of the aforementioned challenges. To mitigate Correspondence Drift, we employ a \textbf{Cross-Task Synergy} training paradigm, co-training the joint generation task with 
\hutnew{auxiliary audio-driven video
and video-driven audio generation tasks} to leverage the latter's strong supervisory signal for instilling robust alignment priors. 
\hut{To resolve the conflation of local and global synchronization objectives,}
we further propose a \textbf{Global-Local Decoupled Interaction Module} that ensures both holistic style consistency and precise temporal synchronization through the decoupled global style attention and localized, \hut{RoPE-aligned} frame-wise attention. 
Finally, to address the fact that conventional CFG lacks a mechanism for enhancing audio-visual alignment, we propose \textbf{Synchronization-Enhanced CFG (SyncCFG)}. This novel technique redefines the negative condition learned from the cross-task training stage to explicitly isolate and amplify the guidance vector corresponding to audio-visual alignment. 

\begin{table}[t]
\centering
\caption{Comparison of capabilities among existing joint audio-video generation models. We evaluate their ability to generate different sound types and the quality of their temporal alignment with the video. (\cmark: Good, \fmark: Fair/Limited, \xmark: Poor/Unsupported)}
\vspace{-0.1in}
\label{tab:related_work_comparison}
\resizebox{\columnwidth}{!}{%
\begin{tabular}{lcccc}
\toprule
\textbf{Model} & \makecell{\textbf{Human} \\ \textbf{Speech}} & \makecell{\textbf{Environmental} \\ \textbf{Sound}} & \makecell{\textbf{Speech-Video} \\ \textbf{Alignment}} & \makecell{\textbf{Sound-Video} \\ \textbf{Alignment}} \\
\midrule
MM-Diffusion~\cite{ruan2023mmdiffusion} & \xmark & \fmark & \xmark & \fmark \\
JavisDiT~\cite{liu2025javisdit} & \xmark & \fmark & \xmark & \fmark \\
AnimateSI~\cite{wang2025animate} & \xmark & \fmark & \xmark & \fmark \\
JAM-Flow~\cite{kwon2025jam} & \cmark & \xmark & \cmark &\xmark \\
UniVerse-1~\cite{wang2025universe} & \cmark & \cmark & \xmark & \fmark \\
Ovi~\cite{low2025ovi} & \cmark & \cmark & \fmark & \fmark \\
\midrule
\textbf{Harmony (Ours)} & \cmark & \cmark & \cmark & \cmark \\
\bottomrule
\end{tabular}%
}
\vspace{-0.18in}
\end{table}


\hutnew{Extensive experiments on our newly proposed \textbf{Harmony-Bench} validate our framework on the challenging task of jointly generating human speech and ambient sounds. Harmony achieves the best audio-video alignment, maintaining fine-grained temporal synchronization in complex acoustic scenes, which confirms the efficacy of our approach. The main contributions of our work are summarized as follows:}

\begin{itemize}
    \item We propose \textbf{Harmony}, a novel \yr{joint} audio-\yr{video} generation framework built upon the principle of \textbf{Cross-Task Synergy} to resolve the fundamental \textit{Correspondence Drift} problem in joint diffusion models.
    \item We design a \textbf{Global-Local Decoupled Interaction Module} that achieves comprehensive alignment in both overall style and fine-grained temporal details.
    \item We propose a novel \textbf{Synchronization-Enhanced CFG (SyncCFG)} that guides the model towards better audio-visual correspondence during inference by using 
    \yr{mute-audio} \hutnew{and static-video} condition as negative guidance.
    \item We establish a new state-of-the-art in audio-visual generation, with extensive experiments validating Harmony's superior performance in cross-modal synchronization.
\end{itemize}

\section{Related Work}
\label{sec:related_work}

\subsection{Video Generation}
The field of video generation has rapidly advanced~\cite{animatediff,hunyuanvideo,wan,hu2025ultragen,xue2025ultravideo}, transitioning from early Generative Adversarial Networks (GANs)~\cite{gan} to the now-dominant diffusion models~\cite{ddpm}. Building on their success in image synthesis~\cite{ldm, sdxl,hu2025improving}, models like AnimateDiff~\cite{animatediff} and SVD~\cite{svd} extended diffusion to the temporal domain. Architectures have also evolved from UNets to more powerful Diffusion Transformers (DiT)~\cite{cogvideox, opensoraplan}, with recent open-source models like HunyuanVideo~\cite{hunyuanvideo} and Wan~\cite{wan} achieving state-of-the-art visual quality, which inspired a lot of downstream video generation methods, like video customization~\cite{hu2025hunyuancustom,hu2025polyvivid}, video editing~\cite{liang2025omniv2v,chen2025ivebench}, and camera control~\cite{wang2024motionctrl,hu2024motionmaster,hu2025high}. However, a critical limitation persists across this body of work: a singular focus on the visual modality. By generating silent videos, these models produce content that feels incomplete and lacks the immersive quality of real-world experiences, underscoring the need for cohesive audio-visual synthesis.


\subsection{Joint Audio-Video Generation}

Recently, a growing body of research has begun to explore the simultaneous generation of audio and video within a single, unified framework~\citep{liu2025javisdit,ruan2023mmdiffusion,low2025ovi,wang2025universe,wang2025av,xing2024seeing,zhang2025uniavgen}. However, most of the early open-source approaches in this domain were restricted to synthesizing coarse environmental sounds and were unable to generate meaningful human speech~\citep{liu2025javisdit,ruan2023mmdiffusion,ishii2024simple}. This limitation began to be addressed by subsequent models like JAM-Flow~\cite{kwon2025jam} and UniAVGen~\cite{zhang2025uniavgen}, which started to incorporate speech-video joint generation. A significant advancement came with models like UniVerse-1~\citep{wang2025universe} and Ovi~\citep{low2025ovi}, which integrated more powerful audio synthesis components to enable the joint generation of both ambient sounds and human vocals. Despite this progress, a key challenge remains in the detailed alignment of the overall soundscape. These models often struggle to cohesively blend human speech with its surrounding environmental audio in a manner that is acoustically and semantically consistent with the visual context, highlighting a remaining gap in creating truly immersive audio-visual experiences.

\section{Method}
\label{sec:method}

 \begin{figure*}[t]
    \centering
    \includegraphics[width=0.85\textwidth]{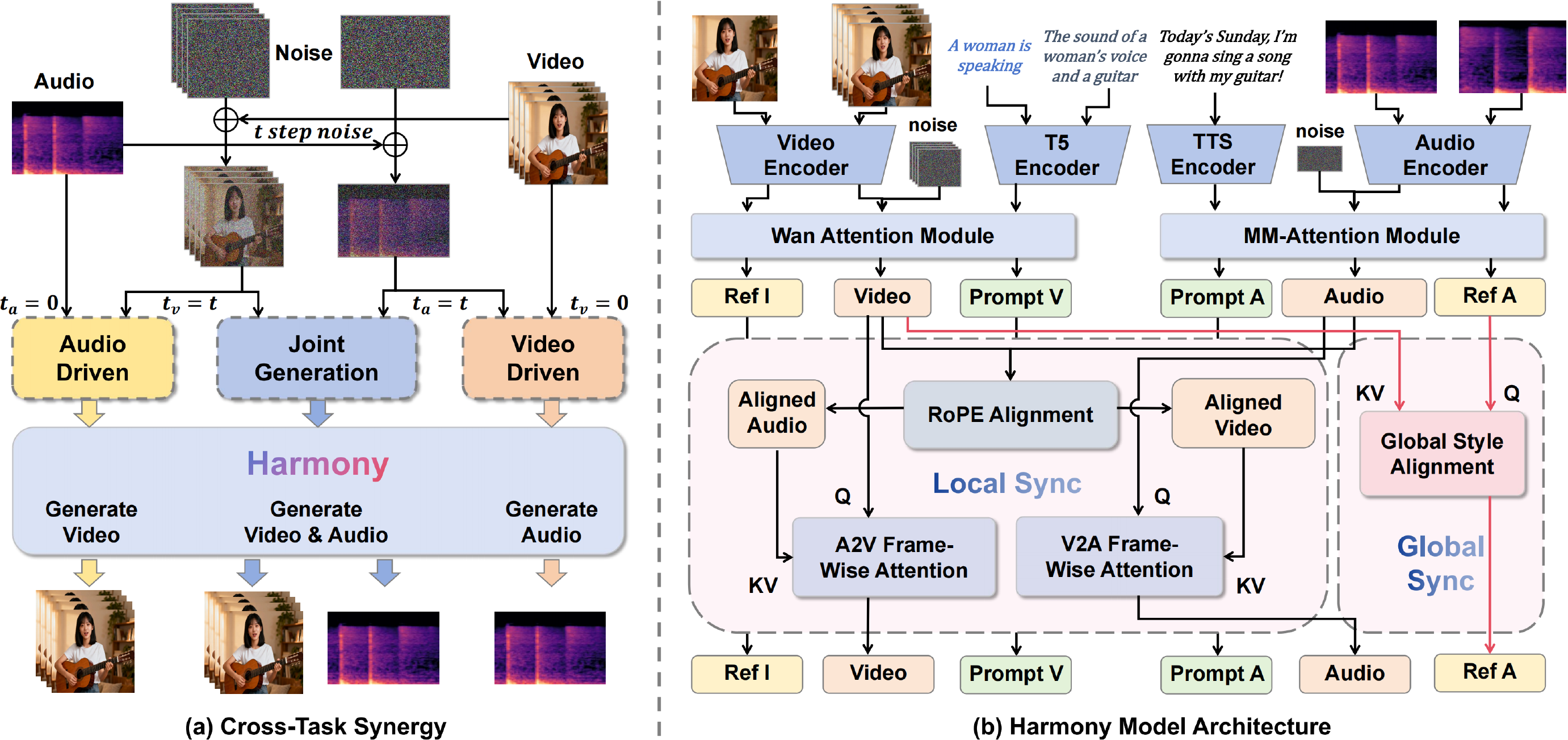}
    \vspace{-0.1in}
    \caption{(a) \textbf{Mitigating Correspondence Drift with Cross-Task Synergy.} Our training paradigm leverages a supervised audio- and video-driven task to provide a strong alignment signal. This instills robust synchronization features in the model, stabilizing the otherwise stochastic joint generation process. (b) \textbf{Overview of the Harmony Model.} The architecture features parallel branches for multimodal inputs. The video stream is conditioned on a reference image and a descriptive prompt. The audio stream is conditioned on a reference audio, an ambient sound description, and a speech transcript. The model then generates a single, synchronized audio-visual result.}
    \vspace{-0.1in}
    \label{fig:framework}
\end{figure*}

In this section, we introduce \textbf{\modelname}, a novel framework for joint audio-\yr{video} synthesis designed to overcome the fundamental challenge of cross-modal alignment in diffusion models\yr{, capable of joint audio-video generation on speech, sound effects, and ambient audio}. 
\yr{We introduce three core innovative designs:}
(1) a \textbf{Cross-Task Synergy} training strategy that \yr{combines the standard joint audio-video generation task with auxiliary 
\hutnew{audio-driven video and video-driven audio generation tasks,}  
}
\hutnew{leveraging the strong, uni-directional supervisory signals from both to accelerate and stabilize the learning of audio-video alignment}
(2) a \textbf{Global-Local Decoupled Interaction Module} that efficiently \yr{ensures} both fine-grained temporal correspondence and holistic stylistic consistency; 
and (3) a \textbf{Cross-Task Alignment-Enhanced CFG} mechanism that repurposes guidance \yr{by designing more meaningful negative anchors,} to explicitly amplify audio-\yr{video} synchronization during inference.

\subsection{Preliminary: Joint Audio-\yr{Video} Diffusion}
\label{subsec:preliminary}

Joint audio-\yr{video} synthesis typically employs a dual-stream Latent Diffusion Model. After encoding a video $V$ and audio $A$ into latents ($z_v, z_a$), a denoising network $\epsilon_\theta$ is trained to reverse a standard Gaussian noising process. The network consists of parallel video and audio backbones that process their respective noisy latents, $z_{v,t}$ and $z_{a,t}$. Synchronization is learned through an interaction module (e.g., cross-attention) that couples the two streams. The model is optimized by minimizing the noise prediction error for both modalities:
\begin{small}
 \begin{equation}
\mathcal{L} = ||\epsilon_v - \hat{\epsilon}_v(z_{v,t}, z_{a,t}, t)||^2 + ||\epsilon_a - \hat{\epsilon}_a(z_{a,t},z_{v,t},  t)||^2.
\end{equation}
\end{small}
However, this standard approach struggles to learn robust alignment from two concurrently noisy signals—a core challenge our work addresses.

\subsection{Cross-Task Synergy for Enhanced Alignment}
\label{subsec:synergy}

 \begin{figure}[t]
    \centering
    \includegraphics[width=0.38\textwidth]{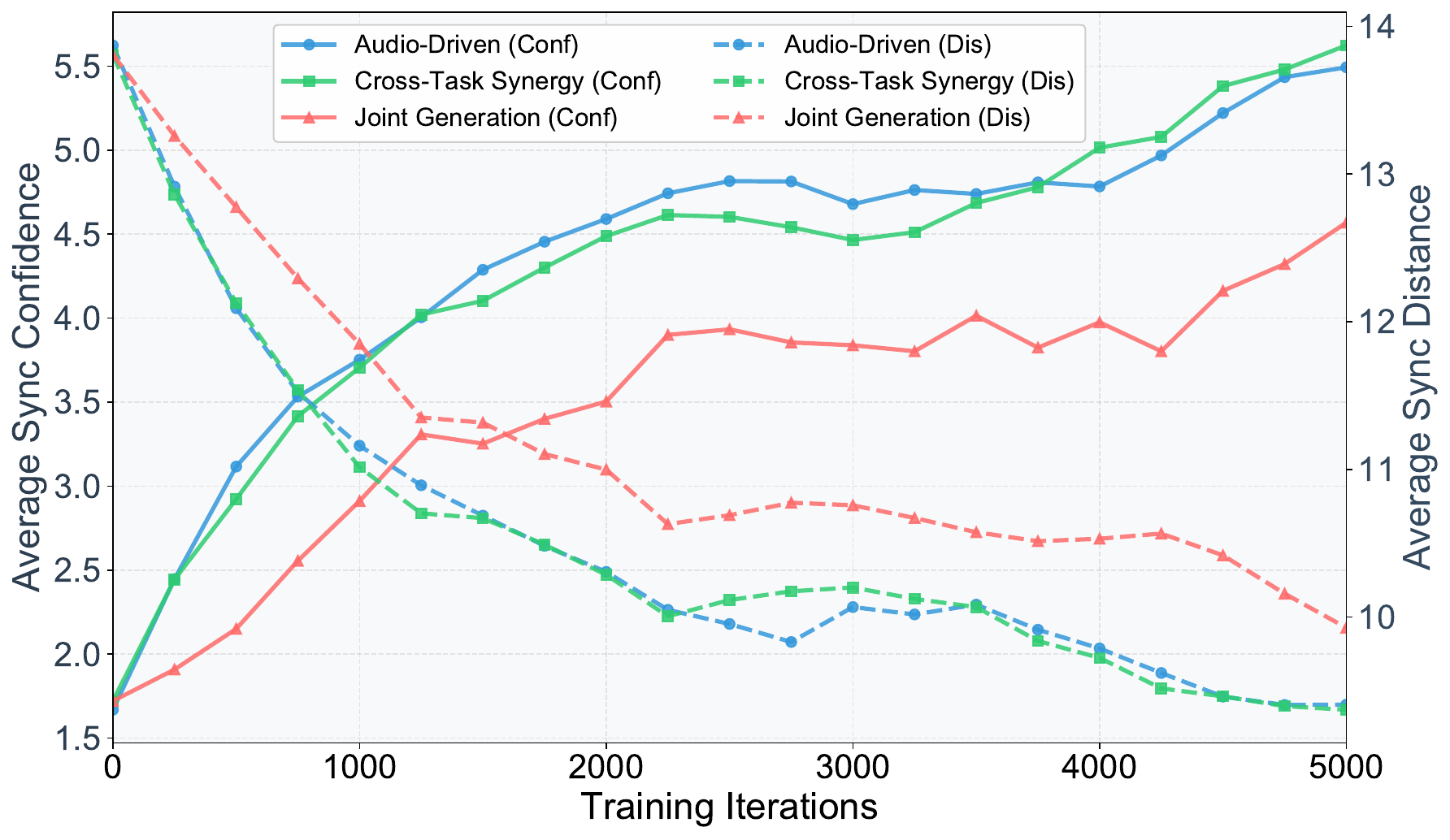}
    \vspace{-0.1in}
    \caption{Comparison of the audio-video alignment score among different training strategies.}
    \vspace{-0.2in}
    \label{fig:motivation}
\end{figure}

\subsubsection{The Challenge of Correspondence Drift}
\label{ssubsec:problem}

\noindent\textbf{Problem Formulation.}
While recent advancements in joint audio-\yr{video} generation have focused on novel architectures, a fundamental challenge persists: achieving robust cross-modal alignment. We identify the root cause not as an architectural limitation, but as an inherent instability in the training paradigm, a phenomenon we term \textbf{Correspondence Drift}. During the initial stages of joint training, both audio and video signals are heavily diffused with noise. Attempting to learn a correspondence between two concurrently evolving, highly stochastic latent variables \yr{results in} an unstable and inefficient learning target, causing the alignment process to drift and converge slowly.

\noindent\textbf{Empirical Motivation.}
To empirically validate this hypothesis, we present a comparative analysis in \cref{fig:motivation}. We \yr{compare} the lip-sync alignment scores of an audio-driven \yr{video generation} task versus a joint \yr{audio-video} generation task, utilizing an identical network architecture (detailed in Sec.~\ref{sec:interaction_module}). The results 
\yr{reveal that}
the audio-driven model, conditioned on a clean audio signal, rapidly converges to a high alignment score; in contrast, the joint generation model exhibits markedly slower convergence. This disparity strongly indicates that anchoring one modality with a deterministic, noise-free signal, \yr{as implemented in the audio-driven model,}
provides a stable learning gradient, enabling the cross-modal interaction module to efficiently capture alignment cues.

\subsubsection{Cross-Task Synergy}
\label{ssubsec:framework}

\noindent\textbf{
\yr{Overview of Cross-Task Synergy.}}
Based on this insight, we propose a novel training framework, \textit{Cross-Task Synergy}, which synergistically combines the standard joint \yr{audio-video} generation task \yr{(the primary task)} with 
\hutnew{audio-driven video and video-driven audio generation tasks.}
By leveraging the high-quality\yr{, noise-free} learning signal from the \hutnew{uni-directional supervisory} task, our model efficiently learns intricate audio-\yr{video} correspondences. This pre-learned alignment knowledge then acts as a powerful catalyst, accelerating the convergence and enhancing the final alignment quality of the primary joint generation task.

\noindent\textbf{Dual-Branch Model Architecture.} 
\hutnew{Our model features a dual-branch architecture for video and audio generation. The video branch adapts the pre-trained Wan2.2-5B model~\cite{wan}. To ensure structural parity, we design a symmetric audio generation branch that synthesizes an audio clip $A$ conditioned on a speech transcript $T_s$ (phonetic content), a descriptive caption $T_a$ (describing the acoustic scene, e.g., vocal emotion or ambient sounds), and a reference audio $A_r$ (timbre). We process these inputs with a multi-encoder setup: an audio VAE~\cite{cheng2025mmaudio} encodes $A$ and $A_r$ into latents $z_a$ and $z_r$. Crucially, departing from prior works~\cite{liu2025javisdit, low2025ovi}, we use separate text encoders to preserve phonetic precision: a dedicated speech-encoder~\cite{chen2024f5} for the transcript $T_s$ ($\to \mathbf{e}_{\text{speech}}$) and a T5 encoder~\cite{chung2023umT5} for the descriptive prompt $T_a$ ($\to \mathbf{e}_{\text{prompt}}$). During denoising, the reference latent $z_r$ is prepended to the noisy target latent $z_{a,t}$, forming a composite input latent $z'_{a,t}$. This composite latent, along with the speech and prompt embeddings, is then processed by a Multi-Modal Diffusion Transformer (MM-DiT) to predict the noise:}
\begin{equation}
    \hat{\epsilon}_a = \text{MM-DiT}(\text{concat}(z'_{a,t}, \mathbf{e}_{\text{speech}}, \mathbf{e}_{\text{prompt}}), t_a).
\end{equation}
To facilitate effective cross-modal \yr{interaction between the two branches}, we instantiate a bidirectional \yr{global-local decoupled} interaction module at each layer, with further details provided in Sec.~\ref{sec:interaction_module}.

\noindent\textbf{Cross-Task Synergy Training.}
\yr{We design a} hybrid training strategy \yr{that} realizes the principle of Cross-Task Synergy. By concurrently training on both \yr{the} joint generation \yr{task} and \hut{and two deterministic, single-modality-driven tasks (audio-\yr{driven} video \yr{gen} and video-\yr{driven} audio \yr{gen})}, we provide the model with a stable alignment signal to counteract Correspondence Drift. 

\hutnew{The audio-driven task conditions video generation on the clean audio latent by setting the audio timestep $t_a$ to 0. Symmetrically, the video-driven task conditions audio generation on the clean video latent by setting the video timestep $t_v$ to 0. The total training objective is a weighted sum of the three corresponding losses:}
\begin{equation}
\label{eq:total_loss}
\mathcal{L} = \mathcal{L}_{\text{joint}} + \lambda_v \mathcal{L}_{\text{driven}}^\text{audio} + \lambda_a \mathcal{L}_{\text{driven}}^\text{video},
\end{equation}
where $\lambda_v$ and $\lambda_a$ are balancing hyperparameters, and $\mathbf{c}$ represents the set of auxiliary conditions (e.g., text prompts and speech embeddings). The loss components are defined as:
\begin{equation}
\begin{aligned}
    \mathcal{L}_{\text{joint}} =& ||\epsilon_v - \hat{\epsilon}_v(z_{v,t}, z_{a,t},\mathbf{c}, t)||^2 \\&+ ||\epsilon_a - \hat{\epsilon}_a(z_{a,t}, z_{v,t}\mathbf{c}, t)||^2, \\
    \mathcal{L}_{\text{driven}}^\text{audio} =& ||\epsilon_v - \hat{\epsilon}_v(z_{v,t}, z_{a,0},\mathbf{c}, t)||^2, \\
    \mathcal{L}_{\text{driven}}^\text{video} =& ||\epsilon_a - \hat{\epsilon}_a(z_{a,t}, z_{v,0}, \mathbf{c}, t)||^2.
\end{aligned}
\end{equation}
This bidirectional, synergistic training approach enables our model to achieve faster convergence and a superior degree of final audio-video alignment.

\subsection{Global-Local Decoupled Interaction Module}
\label{sec:interaction_module}

\hut{Robust audio-\yr{video} synchronization presents a fundamental tension between two objectives: \textbf{\textit{(1)}} precise, fine-grained temporal alignment and \textbf{\textit{(2)}} holistic, global style consistency \yr{(\textit{e.g.}, emotional tone, ambient features)}. Prior works~\cite{liu2025javisdit, low2025ovi} often attempt to address both with a single, monolithic mechanism like global cross-attention, which conflates these goals and leads to a suboptimal trade-off. To resolve this, we propose a novel \textbf{Global-Local Decoupled Interaction Module} with two specialized components: \textbf{\textit{(1)}} a \textbf{\yr{RoPE}-Aligned Frame-wise Attention} module for precise local synchronization, and \textbf{\textit{(2)}} a \textbf{Global Style Alignment} module for holistic consistency. This decoupled design allows each component to excel at its specific task, resolving the conflict between fine-grained temporal alignment and global style propagation.}


\subsubsection{\yr{RoPE}-Aligned Frame-wise Attention}


\hutnew{To achieve precise temporal synchronization, we employ a local frame-wise attention strategy, which is more computationally efficient and better suited for fine-grained alignment than global cross-attention. However, a key challenge arises from the mismatched sampling rates of video and audio latents ($T_v \neq T_a$). This discrepancy means a specific event in one modality can occur at a timepoint that falls \textit{between} two discrete frames in the other. A standard attention mechanism, forced to operate on a discrete set of keys, must attend to the nearest but temporally imperfect frame. This forced approximation introduces temporal jitter and fundamentally degrades fine-grained synchronization.}

\noindent\hut{\textbf{Temporal Alignment via RoPE Scaling.} To resolve this mismatch, we introduce an alignment step prior to the attention operation~\cite{cheng2025mmaudio}. Our key insight is to unify the temporal coordinate spaces of both modalities by dynamically scaling their Rotary Positional Embeddings (RoPE)~\cite{su2024roformer}. Before attention \yr{operation}, we rescale the positional indices of the source modality to match the timeline of the target. For instance, in Audio-to-Video (A2V) attention, an audio frame at index $j$ is mapped to a virtual position $j' = j \cdot (T_v / T_a)$ for its RoPE calculation. This ensures their positional encodings are directly comparable, establishing a strong inductive bias for correct temporal correspondence.}

\noindent\hut{\textbf{Frame-wise Cross-Attention Mechanism.} With the latents now temporally aligned in the RoPE space, we apply a symmetric, bidirectional cross-attention mechanism. 
\yr{Each frame's attention is confined to a small, relevant temporal window in the other modality.}
Taking A2V as an example, given a video latent $z_v$ and an audio latent $z_a$, we first reshape $z_v$ to expose its temporal dimension ($z'_v$). For each video frame $i$, we construct a local context window $C_{a,i}$ from adjacent audio frames. Cross-attention is then applied independently for each video frame, attending to its corresponding audio context window:}
\begin{equation}
\begin{aligned}
    &\Delta z'_v[:, i, :, :] = \text{Cross-Attn}(Q_{v,i}, K_{a,i},V_{a,i}), \forall i \in [0, T_v\text{-}1],\\
    &\yr{Q_{v,i} = z'_v[:, i, :, :] W^Q_{v,i}, K_{a,i} = C_{a,i} W^K_{a,i}, V_{a,i} = C_{a,i} W^V_{a,i}}.
\end{aligned}
\end{equation}
The Video-to-Audio (V2A) \yr{frame-wise} alignment operates analogously. The updates are then integrated via residual connections:
\begin{equation}
\begin{aligned}
    z_v^{\text{updated}} = z_v + \Delta z'_v, \quad
    z_a^{\text{updated}} = z_a +  \Delta z_a.
\end{aligned}
\end{equation}
This RoPE-aligned frame-wise mechanism efficiently enforces mutual temporal synchronization, \yr{leveraging} the benefits of local attention while correctly handling disparate timescales.

\subsubsection{Global Style Alignment}
\hutnew{While frame-wise attention excels at establishing fine-grained temporal correspondence, its localized nature \yr{inherently limits} the propagation of holistic stylistic attributes, such as \yr{the} overall emotional tone or ambient characteristics, \yr{which require a global context to be consistently maintained}. Prior methods often rely on a single, monolithic global attention mechanism, which conflates the distinct tasks of temporal alignment and \yr{global} style \yr{consistency}, overburdening the module. To address this, we propose a principled decoupling: our RoPE-Aligned Frame-wise Attention is exclusively responsible for precise temporal correspondence, while a dedicated \textbf{Global Style Alignment} module handles holistic consistency. This separation allows each component to specialize, preventing interference between the two objectives.}

\hutnew{Our core insight for global alignment is to leverage the reference audio latent, $z_r$ \yr{(which provides speaker identity and timbre)}, as a compact carrier for style information. Instead of directly modifying the target audio $z_a$ and disrupting its fine-grained denoising, we modulate $z_r$ with the global context from the entire video latent $z_v$. This is achieved by treating $z_r$ as the query and $z_v$ as the key and value within a residual cross-attention block:}
\begin{equation}
\begin{aligned}
    &z_r^{\text{updated}} = z_r + \text{Cross-Attn}(Q_r, K_v, V_v), \\
    &\yr{Q_r=z_rW^Q_r, K_v=z_vW^K_v, V_v=z_vW^V_v}.
\end{aligned}
\end{equation}
The resulting visually-informed \yr{reference audio} latent $z_r^{\text{updated}}$ is then prepended to the noisy audio latent $z_{a,t}$ (as described in \cref{ssubsec:framework}), allowing the audio generation to condition on a visually-grounded global style.
This decoupled 
\yr{design offers a key advantage:}
by \yr{confining} the global style injection to the reference latent, \yr{it prevents interference between} holistic \yr{style} consistency \yr{and} 
precise frame-wise temporal alignment, preserving the stability and fidelity of the final audio generation.

 \begin{figure}[t]
    \centering
    \includegraphics[width=0.38\textwidth]{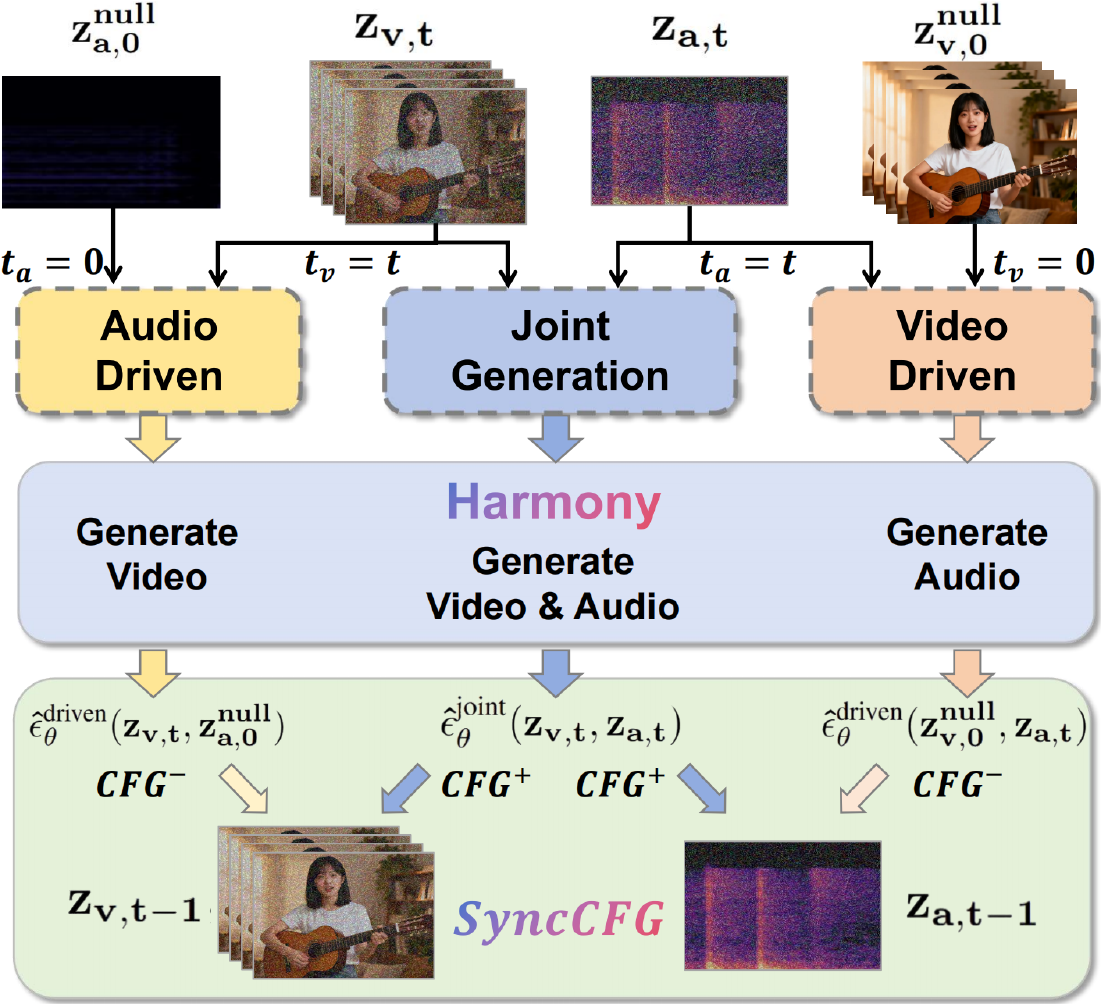}
    \vspace{-0.1in}
    \caption{\textbf{SyncCFG} employs the mute audio and static video as the negative anchors to capture the synchronization feature, which can effectively enhance the audio-video alignment.}
    \vspace{-0.15in}
    \label{fig:SyncCFG}
\end{figure}

\subsection{Synchronization-Enhanced CFG}
\label{subsec:alignment_cfg}

While Classifier-Free Guidance (CFG) is a powerful technique for conditional generation, its standard application in audio-\yr{video} synthesis fails to explicitly amplify the crucial correspondence between modalities. To address this, we introduce \textbf{Synchronization-Enhanced (SyncCFG)}, a novel scheme that repurposes the guidance mechanism to specifically target and enforce audio-\yr{video} synchronization. Our approach leverages the dual capabilities—joint generation and audio\&video-driven synthesis—acquired during our cross-task training to enhance the alignment signal.



\subsubsection{Analysis of Standard Guidance Limitations}
\hutnew{The standard CFG formulation in previous work~\cite{low2025ovi} is used to strengthen the conditioning on a text prompt $c$:}
\begin{small}
\begin{equation}
    \tilde{\epsilon} = \hat{\epsilon}_{\theta}(z_{v,t}, z_{a,t}, \emptyset_c) + s \Big( \hat{\epsilon}_{\theta}(z_{v,t}, z_{a,t}, c) - \hat{\epsilon}_{\theta}(z_{v,t}, z_{a,t}, \emptyset_c) \Big).
    \label{eq:standard_cfg}
\end{equation}
\end{small}
Here, the guidance pushes the denoising process away from an unconditional (null-text $\emptyset_c$) prediction and towards one that aligns with the text prompt $c$.

\hutnew{The key limitation of standard CFG is that its guidance is oriented solely towards text-adherence. The guidance vector, computed by contrasting the text-conditioned output with a text-unconditioned one, exclusively strengthens how well the output matches the prompt. This process, however, is agnostic to the internal consistency between audio and video. It provides no mechanism to isolate or amplify the crucial \textbf{synchronization} signal between the two streams.}

\subsubsection{SyncCFG Formulation for Video Guidance}
\hutnew{To explicitly compute an alignment-enhancing direction, we aim at isolating the visual dynamics caused by the audio. Our key insight is to design a \textit{more meaningful Negative Anchor} that represents a static baseline—how the video should look in the absence of sound. For instance, for a person speaking, the correct video for a silent audio track would be a still face with a closed mouth.}

\hutnew{We achieve this by creating a ``silent audio" negative anchor. We leverage the audio-driven pathway of our model to predict the noise for the video latent $z_{v,t}$ conditioned on a ``muted" audio input, $z_{a,0}^{null}$. The resulting prediction, $\hat{\epsilon}_{\theta}^{\text{driven}}(z_{v,t}, z_{a,0}^{null})$, represents the model's expectation for this visually static scene.}
The guided prediction for the video noise $\tilde{\epsilon}_v$ is then formulated as:
\begin{small}
\begin{equation}
\label{eq:cfg_video}
\begin{aligned}
    \tilde{\epsilon}_v =& \hat{\epsilon}_{\theta}^{\text{driven}}(z_{v,t}, z_{a,0}^{null}) + \\ &s_v \Big( \hat{\epsilon}_{\theta}^{\text{joint}}(z_{v,t}, z_{a,t}) - \hat{\epsilon}_{\theta}^{\text{driven}}(z_{v,t}, z_{a,0}^{null}) \Big).
\end{aligned}
\end{equation}
\end{small}
The subtraction term \hutnew{isolates the precise visual modifications—such as mouth movements or object impacts—that are directly correlated with the audio. By amplifying this difference, SyncCFG specifically enhances the synchronization between sound and motion.}

\begin{figure*}[t]
    \centering
    \includegraphics[width=0.9\textwidth]{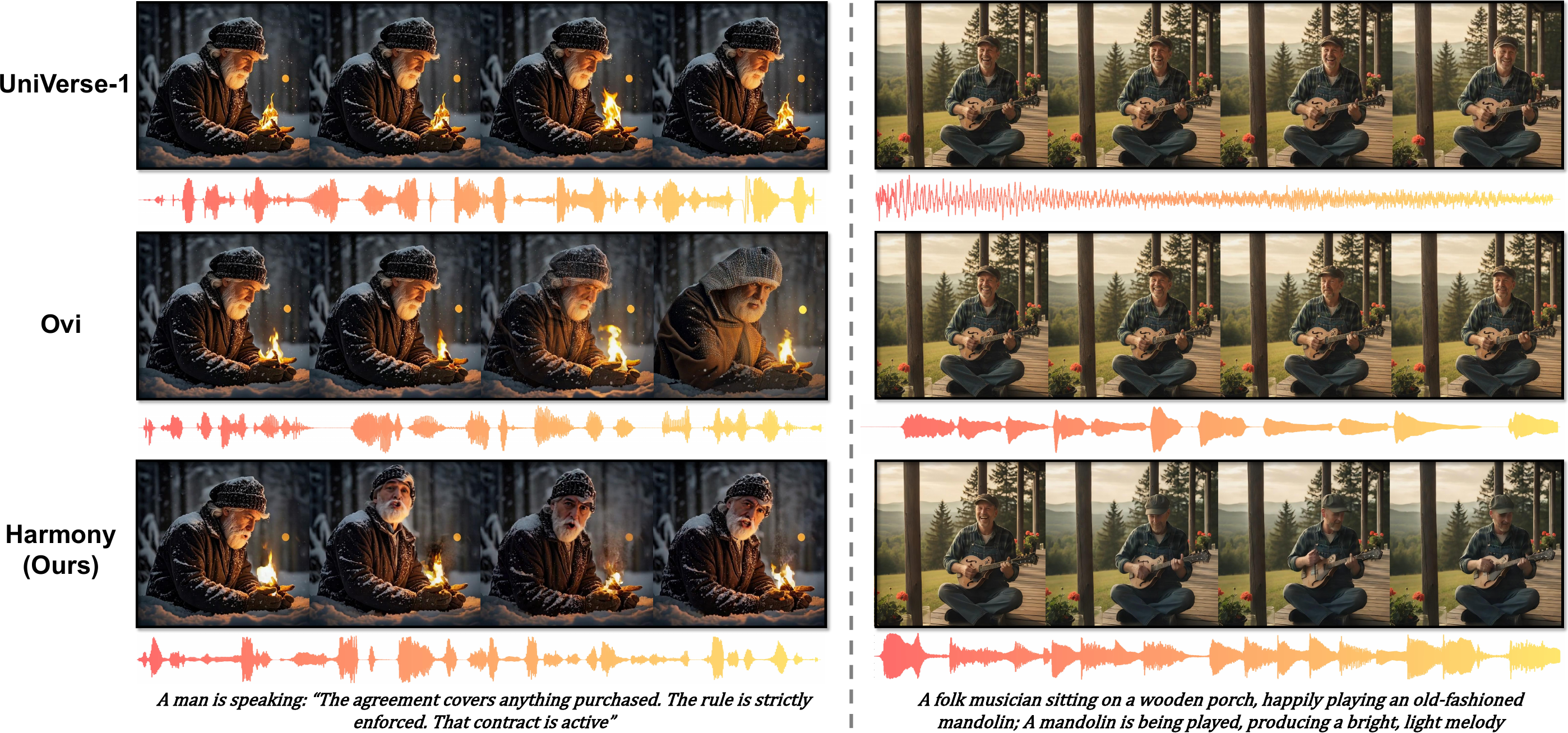}
    \vspace{-0.1in}
    \caption{\textbf{Qualitative Comparison} between Harmony and the state-of-the-art methods, including Universe-1~\cite{wang2025universe} and Ovi~\cite{low2025ovi}.}
    \vspace{-0.1in}
    \label{fig:comparison}
\end{figure*}

\subsubsection{SyncCFG Formulation for Audio Guidance}

\hutnew{\yr{Similarly}, for audio guidance, we \yr{design a null video-based negative anchor} to isolate motion-driven sounds. Using a ``static video" latent $z_{v,0}^{null}$, we predict a baseline audio signal $\hat{\epsilon}_{\theta}^{\text{driven}}(z_{v,0}^{null}, z_{a,t})$ that represents the ambient sound of a motionless scene, \yr{as the video content is static}. The guided \yr{prediction for} audio noise $\tilde{\epsilon}_a$ is \yr{then formulated as}:}
\begin{small}
\begin{equation}
\label{eq:cfg_audio}
\begin{aligned}
    \tilde{\epsilon}_a = &\hat{\epsilon}_{\theta}^{\text{driven}}(z_{v,0}^{null}, z_{a,t}) +\\& s_a \Big(  \hat{\epsilon}_{\theta}^{\text{joint}}(z_{v,t}, z_{a,t}) - \hat{\epsilon}_{\theta}^{\text{driven}}(z_{v,0}^{null}, z_{a,t}) \Big).
\end{aligned}
\end{equation}
\end{small}
This approach transforms CFG from a generic conditional amplifier into a targeted mechanism, \yr{effectively} enforcing fine-grained audio-\yr{video} correspondence.



\section{Experiments}
\label{sec:experiment}

\begin{table*}[t]
\centering
\small
\caption{\textbf{Quantitative comparison} with state-of-the-art methods averaging on ambient sound-video, speech-video, and complex scene-video generation. We evaluate performance across three categories: video quality, audio fidelity, and audio-visual synchronization. Best results are in \textbf{bold}, second-best are \underline{underlined}. For more comprehensive and detailed evaluations, please refer to the supplementary material. Some metrics could not be generated for MM-Diffusion, as it is an unconditional generation model.}
\label{tab:joint_comparison}
\vspace{-0.1in}
\resizebox{\linewidth}{!}{
    \renewcommand{\arraystretch}{1.2}
    \begin{tabular}{l|ccccc|cccccc|cccc}
    \toprule
    \multirow{2}{*}{Method} & \multicolumn{5}{c|}{Video Quality \& Coherence} & \multicolumn{6}{c|}{Audio Fidelity \& Quality} & \multicolumn{4}{c}{Audio-Visual Synchronization} \\
    \cmidrule(lr){2-6} \cmidrule(lr){7-12} \cmidrule(lr){13-16}
     & AQ$~\uparrow$ & IQ$~\uparrow$ & DD$~\uparrow$ & MS$~\uparrow$ & ID$~\uparrow$ &  PQ$~\uparrow$ & PC$~\downarrow$ & CE$~\uparrow$ & CU$~\uparrow$ & WER$~\downarrow$ & IB-A$~\uparrow$ & Sync-C$~\uparrow$ & Sync-D$~\downarrow$ & DeSync$~\downarrow$ &  IB$~\uparrow$ \\
    \midrule

     MM-Diffusion~\cite{ruan2023mmdiffusion} & 0.32 & 0.43 & 0.13 & 0.99  & - & 5.37 & 4.07 & 4.27 & \textbf{5.89} & - & - & - & - & - & 0.12 \\
    
    JavisDiT~\cite{liu2025javisdit} & 0.34 & 0.53 & \textbf{0.38} & 0.99 & 0.38  & 5.46 & 2.24 & 3.19 & 4.54 & 1.00 & \textbf{0.14} & 0.89 & 11.62 & 1.13 & \underline{0.18} \\

    UniVerse-1~\cite{wang2025universe} & 0.52 & \textbf{0.67} & 0.24 & 0.99 & 0.89  & 5.52 & \underline{2.13} & 3.63 & 4.84 & \underline{0.24} & 0.07 & 0.97 & 10.71 & \underline{1.10} & 0.12 \\

    Ovi~\cite{low2025ovi}  & \underline{0.57} & \underline{0.65} & 0.34 & 0.99 & \underline{0.90}  & \underline{6.19} & \underline{2.13} & \underline{4.44} & \underline{5.84} & 0.49 & \underline{0.12} & \underline{4.04} & \underline{9.62} & 1.14 & \underline{0.18} \\
      
    \rowcolor{CornflowerBlue!20}
    \textbf{Harmony (Ours)} & \textbf{0.59} & \underline{0.65} & \underline{0.36} & 0.99 & \textbf{0.91}  & \textbf{6.39} & \textbf{2.05} & \textbf{4.73} & 5.67 & \textbf{0.15} & \underline{0.12} & \textbf{5.61} & \textbf{7.53} & \textbf{0.92} & \textbf{0.19}\\
    
    \bottomrule
    \end{tabular}
}
\vspace{-0.1in}
\end{table*}

\subsection{Experimental Settings}

\noindent\textbf{Datasets and Training.}
Our model is trained on a diverse corpus of over 4 million audio-visual clips, covering both human speech and environmental sounds. The data is aggregated from public sources like OpenHumanVid~\cite{li2025openhumanvid}, AudioCaps~\cite{kim2019audiocaps}, and WavCaps~\cite{mei2024wavcaps}, and supplemented with our own curated high-quality collections. All data is uniformly annotated using Gemini~\cite{gemini_website}. Our training follows a three-stage curriculum: (1) foundational audio pre-training on all audio data, (2) timbre disentanglement finetuning using multi-utterance speech data, and (3) final cross-task joint audio-visual training. The video branch is initialized from Wan2.2~\cite{wan}. The final joint stage is trained for 10,000 iterations with a batch size of 128 and a learning rate of 1e-5. A comprehensive list of datasets, data processing details, and full training hyperparameters are provided in the supplementary material.

\noindent\textbf{Harmony-Bench and Metrics.}
To facilitate a rigorous evaluation, we introduce \textbf{Harmony-Bench}, a new benchmark of 150 test cases designed to assess core audio-visual generation capabilities. It is structured into three 50-item subsets with increasing complexity:
\begin{enumerate}
    \item \textbf{Ambient Sound-Video:} Evaluates temporal alignment of non-speech sounds using AI-generated scenarios conditioned on audio and video captions.
    \item \textbf{Speech-Video:} Assesses lip-sync and speech quality on a mix of real-world and synthetic multilingual data, conditioned primarily on a transcript.
    \item \textbf{Complex Scene (Ambient + Speech):} Tests the model's ability to generate and synchronize co-occurring speech and ambient sounds in complex scenes, using a full set of multimodal prompts.
\end{enumerate}
We evaluate performance using a comprehensive suite of automated metrics. \textbf{For Video}, we measure \textit{Aesthetic Quality} (aesthetic-predictor-v2-5), \textit{Imaging Quality} (MUSIQ), \textit{Dynamic Degree} (RAFT), \textit{Motion Smoothness} and \textit{Identity Consistency} (DINOv3). \textbf{For Audio}, we report \textit{AudioBox-Aesthetics} (PQ, PC, CE, CU), \textit{WER} (Whisper-large-v3), and \textit{IB-A Score}. \textbf{For AV-Sync}, we use lip-sync metrics (\textit{Sync-C}, \textit{Sync-D}), 
and overall consistency (\textit{IB-score}). Further details on the benchmark construction and metric implementations are available in the appendix.

\subsection{Comparison on audio-video generation}
To evaluate the performance of our model, we compare it with state-of-the-art audio-video generation methods on the three types of datasets (Ambient Sound-Video, Speech-Video, and Complex Scene), including MM-Diffusion~\cite{ruan2023mmdiffusion}, JavisDiT~\cite{liu2025javisdit}, UniVerse-1~\cite{wang2025universe}, and Ovi~\cite{low2025ovi}. The quantitative results are shown in Tab.~\ref{tab:joint_comparison}. Our model, Harmony, demonstrates a highly competitive performance, achieving state-of-the-art or comparable results in both video quality (e.g., AQ, DD, ID) and audio fidelity (e.g., PC, PQ). Most notably, its primary advantage lies in audio-visual synchronization. Harmony significantly outperforms all baselines on key synchronization metrics, achieving the highest Sync-C score of \textbf{5.61} and the lowest (best) Sync-D score of \textbf{7.53}. This substantial improvement in temporal alignment directly validates the effectiveness of our proposed cross-task synergy mechanism in enhancing cross-modal coherence.

We provide qualitative comparisons with UniVerse-1 and Ovi in Fig.~\ref{fig:comparison}. 
In the talking head example (left), both competing methods fail to produce synchronized lip movements. 
For the music-driven case (right), their limitations persist: UniVerse-1 generates irrelevant noise, while Ovi produces audio that, while musically correct, is less dynamic—a fact reflected in its simpler waveform. 
Visually, both methods yield videos with minimal motion. 
In contrast, our Harmony generates a fluid video of a person playing the mandolin with motions that are dynamically synchronized with the rich, corresponding music, as evidenced by the more complex audio waveform.

 \begin{figure}[t]
    \centering
    \includegraphics[width=0.42\textwidth]{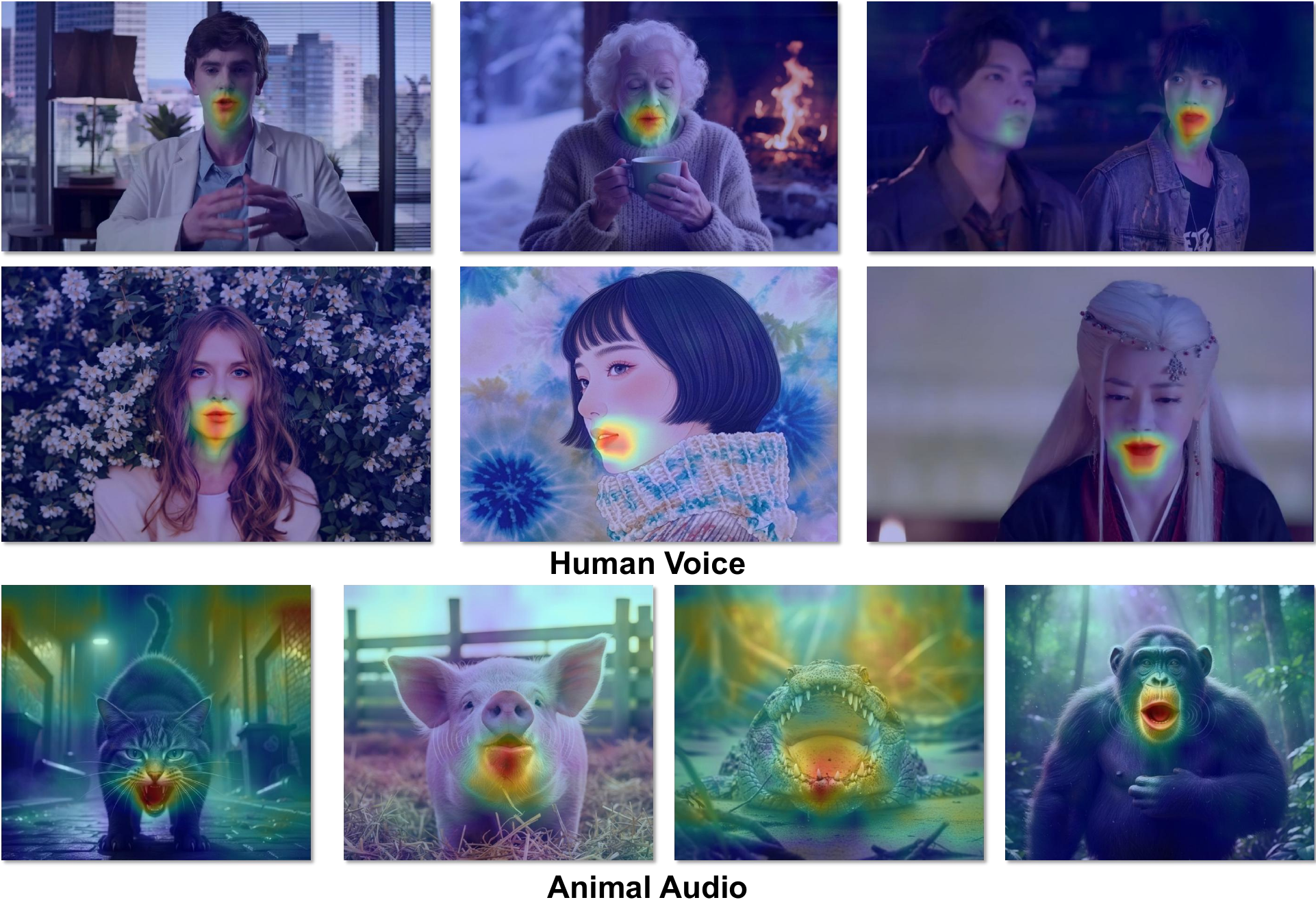}
    \vspace{-0.1in}
    \caption{Visualization of the audio-to-video frame-wise cross-attention map, where the audio can accurately capture the sound source from the videos.}
    \vspace{-0.15in}
    \label{fig:attention map}
\end{figure}

\subsection{Visualization of Cross-modal Attention}

To validate the effectiveness of our frame-wise cross-attention mechanism, we visualize the attention maps from the audio-to-video module. 
As illustrated in Fig.~\ref{fig:attention map}, when synthesizing human speech, the model precisely localizes its attention on the speaker's oral region. 
Notably, in scenarios with multiple individuals, our model can distinguish between them, focusing exclusively on the active speaker. 
This capability extends to natural sounds, where the model accurately identifies the primary sound source (e.g., an animal) while also attending to ambient environmental sounds, such as the rain in the cat example and the birdsong in the crocodile case. 
Collectively, these visualizations underscore our model's superior ability to achieve fine-grained and contextually aware audio-visual alignment.

\subsection{Ablation Studies}

We conduct a comprehensive ablation study to validate our core components, with the results presented in Tab.~\ref{tab:ablation}. In this study, we train all ablated models on the human speech dataset and evaluate their audio-visual alignment. Our baseline model replaces the proposed Global-Local Decoupled Interaction (GLDI) module with a standard global cross-attention mechanism, similar to Ovi~\cite{low2025ovi}, and is trained without cross-task synergy (CTS). As shown in the table, progressively integrating our contributions yields consistent improvements. First, introducing the GLDI module demonstrates the benefit of decoupling local and global interactions. This is further enhanced by the RoPE Alignment, which effectively resolves timescale mismatches and boosts fine-grained synchronization (Sync-C from 4.29 to 4.80). Subsequently, the Cross-Task Synergy (CTS) training strategy further refines the model's alignment capabilities. Finally, applying the Synchronization-Enhanced CFG (SyncCFG) during inference provides the most substantial performance gain, catapulting Sync-C from 5.09 to \textbf{6.51}. This systematic improvement validates that each component of Harmony is crucial for achieving the state-of-the-art audio-video synchronization performance.

\begin{table}[t]
\centering
\caption{Ablation study on the core components of Harmony. We start with a baseline and progressively add each module: Global-local decoupled interaction module (GLDI), RoPE Alignment (RoPE), Cross-Task Synergy (CTS) , and Synchronization-Enhanced CFG (SyncCFG). Note that this experiment is evaluated on the human-speech dataset; therefore, it has different synchronization results compared to Tab.~\ref{tab:joint_comparison}.}
\vspace{-0.05in}
\label{tab:ablation}
\resizebox{0.95\columnwidth}{!}{%
\begin{tabular}{cccc|ccc}
\toprule
\multicolumn{2}{c}{Model Structure}&\multicolumn{2}{c}{Methodology}
& \multicolumn{3}{c}{Synchronization Metrics} \\ 
\cmidrule(lr){1-4} \cmidrule(lr){5-7}
GLDI  &RoPE &  CTS & SyncCFG &  Sync-C$~\uparrow$ & Sync-D$~\downarrow$ &  IB$~\uparrow$ \\
\midrule
  &   &   & & 4.20 & 10.93 & 0.13  \\
\checkmark &   & &   & 4.29 & 10.67  & 0.14 \\
\checkmark  & \checkmark &  &   & 4.80 & 10.30  & 0.14 \\
\checkmark  & \checkmark & \checkmark & & 5.09 & 10.16 & 0.15  \\
\rowcolor{CornflowerBlue!20} \checkmark  & \checkmark & \checkmark & \checkmark    & \textbf{6.51} & \textbf{8.63} & \textbf{0.18}\\

\bottomrule
\end{tabular}%
}
\vspace{-0.15in}
\end{table}

\section{Conclusion}
\label{sec:conclusion}

In this work, we presented Harmony, a novel framework addressing the audio-visual synchronization gap in generative models. We find this gap stems from key methodological flaws: Correspondence Drift, an architectural conflict between global style and local timing, and the limitations of standard CFG for cross-modal alignment.
To address these issues, Harmony introduces three core components. Cross-Task Synergy training instills robust alignment priors to counteract drift. A Global-Local Decoupled Module resolves the architectural conflict by handling style and timing separately. Finally, our novel Synchronization-Enhanced CFG (SyncCFG) provides an explicit mechanism to amplify the alignment signal during inference.
Our experiments validate that Harmony establishes a new state-of-the-art in audio-video synchronization, proving more effective than simply scaling up models. We believe this work provides a strong foundation for a new generation of accessible and well-aligned audio-visual models.

\section*{Acknowledgments}
\label{sec:ack}
This work is supported by Tencent Hunyuan. We thank our team lead, Yuan Zhou, and Qinglin Lu, for their guidance. We also thank Zhentao Yu, Guozheng Zhang, Zhengguang Zhou, Youliang Zhang, Yi Chen, Zixiang Zhou, and Sen Liang for their help with data and technique support. 



{
    \small
    \bibliographystyle{ieeenat_fullname}
    \bibliography{main}

@String(CVPR= {IEEE Conf. Comput. Vis. Pattern Recog.})

@String(ICASSP= {ICASSP})

@String(ICLR = {Int. Conf. Learn. Represent.})

@String(CVPR  = {CVPR})

@String(ICLR  = {ICLR})

@article{haji2024av,
  title={AV-Link: Temporally-Aligned Diffusion Features for Cross-Modal Audio-Video Generation},
  author={Haji-Ali, Moayed and Menapace, Willi and Siarohin, Aliaksandr and Skorokhodov, Ivan and Canberk, Alper and Lee, Kwot Sin and Ordonez, Vicente and Tulyakov, Sergey},
  journal={arXiv preprint arXiv:2412.15191},
  year={2024}
}

@article{ishii2024simple,
  title={A simple but strong baseline for sounding video generation: Effective adaptation of audio and video diffusion models for joint generation},
  author={Ishii, Masato and Hayakawa, Akio and Shibuya, Takashi and Mitsufuji, Yuki},
  journal={arXiv preprint arXiv:2409.17550},
  year={2024}
}

@inproceedings{ruan2023mmdiffusion,
  title={Mm-diffusion: Learning multi-modal diffusion models for joint audio and video generation},
  author={Ruan, Ludan and Ma, Yiyang and Yang, Huan and He, Huiguo and Liu, Bei and Fu, Jianlong and Yuan, Nicholas Jing and Jin, Qin and Guo, Baining},
  booktitle={Proceedings of the IEEE/CVF Conference on Computer Vision and Pattern Recognition},
  pages={10219--10228},
  year={2023}
}

@article{chung2023umT5,
  title={Unimax: Fairer and more effective language sampling for large-scale multilingual pretraining},
  author={Chung, Hyung Won and Constant, Noah and Garcia, Xavier and Roberts, Adam and Tay, Yi and Narang, Sharan and Firat, Orhan},
  journal={arXiv preprint arXiv:2304.09151},
  year={2023}
}

@misc{veo3_website,
  author       = {Google DeepMind},  
  title        = {Veo3},  
  howpublished = {\url{https://deepmind.google/models/veo/}},
   year         = {2025},
}

@misc{sora2_website,
  author       = {Openai},  
  title        = {Sora2},  
  howpublished = {\url{https://openai.com/zh-Hans-CN/index/sora-2/}},
   year         = {2025},
}

@misc{gemini_website,
  author       = {Google},  
  title        = {Gemini},  
  howpublished = {\url{https://gemini.google.com/}},
   year         = {2025},
}

@article{hu2025hunyuancustom,
  title={HunyuanCustom: A Multimodal-Driven Architecture for Customized Video Generation},
  author={Hu, Teng and Yu, Zhentao and Zhou, Zhengguang and Liang, Sen and Zhou, Yuan and Lin, Qin and Lu, Qinglin},
  journal={arXiv preprint arXiv:2505.04512},
  year={2025}
}

@article{hu2024motionmaster,
  title={Motionmaster: Training-free camera motion transfer for video generation},
  author={Hu, Teng and Zhang, Jiangning and Yi, Ran and Wang, Yating and Huang, Hongrui and Weng, Jieyu and Wang, Yabiao and Ma, Lizhuang},
  journal={arXiv preprint arXiv:2404.15789},
  year={2024}
}

@article{chen2024f5,
  title={F5-tts: A fairytaler that fakes fluent and faithful speech with flow matching},
  author={Chen, Yushen and Niu, Zhikang and Ma, Ziyang and Deng, Keqi and Wang, Chunhui and Zhao, Jian and Yu, Kai and Chen, Xie},
  journal={arXiv preprint arXiv:2410.06885},
  year={2024}
}

@article{shan2025hunyuanfoley,
  title={Hunyuanvideo-foley: Multimodal diffusion with representation alignment for high-fidelity foley audio generation},
  author={Shan, Sizhe and Li, Qiulin and Cui, Yutao and Yang, Miles and Wang, Yuehai and Yang, Qun and Zhou, Jin and Zhong, Zhao},
  journal={arXiv preprint arXiv:2508.16930},
  year={2025}
}

@inproceedings{cheng2025mmaudio,
  title={MMAudio: Taming Multimodal Joint Training for High-Quality Video-to-Audio Synthesis},
  author={Cheng, Ho Kei and Ishii, Masato and Hayakawa, Akio and Shibuya, Takashi and Schwing, Alexander and Mitsufuji, Yuki},
  booktitle={Proceedings of the Computer Vision and Pattern Recognition Conference},
  pages={28901--28911},
  year={2025}
}

@article{peng2025omnisync,
  title={Omnisync: Towards universal lip synchronization via diffusion transformers},
  author={Peng, Ziqiao and Liu, Jiwen and Zhang, Haoxian and Liu, Xiaoqiang and Tang, Songlin and Wan, Pengfei and Zhang, Di and Liu, Hongyan and He, Jun},
  journal={arXiv preprint arXiv:2505.21448},
  year={2025}
}

@inproceedings{xing2024seeing,
  title={Seeing and hearing: Open-domain visual-audio generation with diffusion latent aligners},
  author={Xing, Yazhou and He, Yingqing and Tian, Zeyue and Wang, Xintao and Chen, Qifeng},
  booktitle={Proceedings of the IEEE/CVF Conference on Computer Vision and Pattern Recognition},
  pages={7151--7161},
  year={2024}
}

@inproceedings{wang2025av,
  title={AV-DiT: Taming Image Diffusion Transformers for Efficient Joint Audio and Video Generation},
  author={Wang, Kai and Deng, Shijian and Shi, Jing and Hatzinakos, Dimitrios and Tian, Yapeng},
  booktitle={Proceedings of the 33rd ACM International Conference on Multimedia},
  pages={10486--10495},
  year={2025}
}

@article{wang2025universe,
  title={UniVerse-1: Unified Audio-Video Generation via Stitching of Experts},
  author={Wang, Duomin and Zuo, Wei and Li, Aojie and Chen, Ling-Hao and Liao, Xinyao and Zhou, Deyu and Yin, Zixin and Dai, Xili and Jiang, Daxin and Yu, Gang},
  journal={arXiv preprint arXiv:2509.06155},
  year={2025}
}

@article{low2025ovi,
  title={Ovi: Twin Backbone Cross-Modal Fusion for Audio-Video Generation},
  author={Low, Chetwin and Wang, Weimin and Katyal, Calder},
  journal={arXiv preprint arXiv:2510.01284},
  year={2025}
}

@article{zhang2025speakervid,
  title={SpeakerVid-5M: A Large-Scale High-Quality Dataset for Audio-Visual Dyadic Interactive Human Generation},
  author={Zhang, Youliang and Li, Zhaoyang and Wang, Duomin and Zhang, Jiahe and Zhou, Deyu and Yin, Zixin and Dai, Xili and Yu, Gang and Li, Xiu},
  journal={arXiv preprint arXiv:2507.09862},
  year={2025}
}

@article{ho2022classifier,
  title={Classifier-free diffusion guidance},
  author={Ho, Jonathan and Salimans, Tim},
  journal={arXiv preprint arXiv:2207.12598},
  year={2022}
}

@article{liu2025javisdit,
  title={Javisdit: Joint audio-video diffusion transformer with hierarchical spatio-temporal prior synchronization},
  author={Liu, Kai and Li, Wei and Chen, Lai and Wu, Shengqiong and Zheng, Yanhao and Ji, Jiayi and Zhou, Fan and Jiang, Rongxin and Luo, Jiebo and Fei, Hao and others},
  journal={arXiv preprint arXiv:2503.23377},
  year={2025}
}

@inproceedings{animatediff, 
  title={AnimateDiff: Animate Your Personalized Text-to-Image Diffusion Models without Specific Tuning},
  author={Yuwei Guo and Ceyuan Yang and Anyi Rao and Zhengyang Liang and Yaohui Wang and Yu Qiao and Maneesh Agrawala and Dahua Lin and Bo Dai},
  booktitle={ICLR},
  year={2024}
}

@inproceedings{gan, 
  title = {Generative Adversarial Nets},
  author={Goodfellow, Ian J. and Pouget-Abadie, Jean and Mirza, Mehdi and Xu, Bing and Warde-Farley, David and Ozair, Sherjil and Courville, Aaron and Bengio, Yoshua},
  booktitle={NeurIPS},
  year={2014}
}

@article{ddpm,
  title={Denoising diffusion probabilistic models},
  author={Ho, Jonathan and Jain, Ajay and Abbeel, Pieter},
  journal={NeurIPS},
  year={2020}
}

@inproceedings{ldm,
  title={High-resolution image synthesis with latent diffusion models},
  author={Rombach, Robin and Blattmann, Andreas and Lorenz, Dominik and Esser, Patrick and Ommer, Bj{\"o}rn},
  booktitle={CVPR},
  year={2022}
}

@inproceedings{sdxl,
  title={{SDXL}: Improving Latent Diffusion Models for High-Resolution Image Synthesis},
  author={Dustin Podell and Zion English and Kyle Lacey and Andreas Blattmann and Tim Dockhorn and Jonas M{\"u}ller and Joe Penna and Robin Rombach},
  booktitle={ICLR},
  year={2024}
}

@article{cogvideox,
  title={Cogvideox: Text-to-video diffusion models with an expert transformer},
  author={Yang, Zhuoyi and Teng, Jiayan and Zheng, Wendi and Ding, Ming and Huang, Shiyu and Xu, Jiazheng and Yang, Yuanming and Hong, Wenyi and Zhang, Xiaohan and Feng, Guanyu and others},
  journal={arXiv preprint arXiv:2408.06072},
  year={2024}
}

@article{opensoraplan,
  title={Open-Sora Plan: Open-Source Large Video Generation Model},
  author={Lin, Bin and Ge, Yunyang and Cheng, Xinhua and Li, Zongjian and Zhu, Bin and Wang, Shaodong and He, Xianyi and Ye, Yang and Yuan, Shenghai and Chen, Liuhan and others},
  journal={arXiv preprint arXiv:2412.00131},
  year={2024}
}

@article{hunyuanvideo,
  title={Hunyuanvideo: A systematic framework for large video generative models},
  author={Kong, Weijie and Tian, Qi and Zhang, Zijian and Min, Rox and Dai, Zuozhuo and Zhou, Jin and Xiong, Jiangfeng and Li, Xin and Wu, Bo and Zhang, Jianwei and others},
  journal={arXiv preprint arXiv:2412.03603},
  year={2024}
}

@article{wan,
  title={Wan: Open and Advanced Large-Scale Video Generative Models}, 
  author={Ang Wang and Baole Ai and Bin Wen, et al},
  journal = {arXiv preprint arXiv:2503.20314},
  year={2025}
}

@article{svd,
  title={Stable video diffusion: Scaling latent video diffusion models to large datasets},
  author={Blattmann, Andreas and Dockhorn, Tim and Kulal, Sumith and Mendelevitch, Daniel and Kilian, Maciej and Lorenz, Dominik and Levi, Yam and English, Zion and Voleti, Vikram and Letts, Adam and others},
  journal={arXiv preprint arXiv:2311.15127},
  year={2023}
}

@inproceedings{kim2019audiocaps,
  title={Audiocaps: Generating captions for audios in the wild},
  author={Kim, Chris Dongjoo and Kim, Byeongchang and Lee, Hyunmin and Kim, Gunhee},
  booktitle={Proceedings of the 2019 Conference of the North American Chapter of the Association for Computational Linguistics: Human Language Technologies, Volume 1 (Long and Short Papers)},
  pages={119--132},
  year={2019}
}

@inproceedings{drossos2020clotho,
  title={Clotho: An audio captioning dataset},
  author={Drossos, Konstantinos and Lipping, Samuel and Virtanen, Tuomas},
  booktitle={ICASSP 2020-2020 IEEE International Conference on Acoustics, Speech and Signal Processing (ICASSP)},
  pages={736--740},
  year={2020},
  organization={IEEE}
}

@article{mei2024wavcaps,
  title={Wavcaps: A chatgpt-assisted weakly-labelled audio captioning dataset for audio-language multimodal research},
  author={Mei, Xinhao and Meng, Chutong and Liu, Haohe and Kong, Qiuqiang and Ko, Tom and Zhao, Chengqi and Plumbley, Mark D and Zou, Yuexian and Wang, Wenwu},
  journal={IEEE/ACM Transactions on Audio, Speech, and Language Processing},
  volume={32},
  pages={3339--3354},
  year={2024},
  publisher={IEEE}
}

@inproceedings{chen2020vggsound,
  title={Vggsound: A large-scale audio-visual dataset},
  author={Chen, Honglie and Xie, Weidi and Vedaldi, Andrea and Zisserman, Andrew},
  booktitle={ICASSP 2020-2020 IEEE International Conference on Acoustics, Speech and Signal Processing (ICASSP)},
  pages={721--725},
  year={2020},
  organization={IEEE}
}

@inproceedings{he2024emilia,
  title={Emilia: An extensive, multilingual, and diverse speech dataset for large-scale speech generation},
  author={He, Haorui and Shang, Zengqiang and Wang, Chaoren and Li, Xuyuan and Gu, Yicheng and Hua, Hua and Liu, Liwei and Yang, Chen and Li, Jiaqi and Shi, Peiyang and others},
  booktitle={2024 IEEE Spoken Language Technology Workshop (SLT)},
  pages={885--890},
  year={2024},
  organization={IEEE}
}

@inproceedings{li2025openhumanvid,
  title={Openhumanvid: A large-scale high-quality dataset for enhancing human-centric video generation},
  author={Li, Hui and Xu, Mingwang and Zhan, Yun and Mu, Shan and Li, Jiaye and Cheng, Kaihui and Chen, Yuxuan and Chen, Tan and Ye, Mao and Wang, Jingdong and others},
  booktitle={Proceedings of the Computer Vision and Pattern Recognition Conference},
  pages={7752--7762},
  year={2025}
}

@article{su2024roformer,
  title={Roformer: Enhanced transformer with rotary position embedding},
  author={Su, Jianlin and Ahmed, Murtadha and Lu, Yu and Pan, Shengfeng and Bo, Wen and Liu, Yunfeng},
  journal={Neurocomputing},
  volume={568},
  pages={127063},
  year={2024},
  publisher={Elsevier}
}

@article{zhang2025uniavgen,
  title={UniAVGen: Unified Audio and Video Generation with Asymmetric Cross-Modal Interactions},
  author={Zhang, Guozhen and Zhou, Zixiang and Hu, Teng and Peng, Ziqiao and Zhang, Youliang and Chen, Yi and Zhou, Yuan and Lu, Qinglin and Wang, Limin},
  journal={arXiv preprint arXiv:2511.03334},
  year={2025}
}

@article{kwon2025jam,
  title={JAM-Flow: Joint Audio-Motion Synthesis with Flow Matching},
  author={Kwon, Mingi and Shin, Joonghyuk and Jung, Jaeseok and Park, Jaesik and Uh, Youngjung},
  journal={arXiv preprint arXiv:2506.23552},
  year={2025}
}

@inproceedings{wang2025animate,
  title={Animate and Sound an Image},
  author={Wang, Xihua and Song, Ruihua and Li, Chongxuan and Cheng, Xin and Li, Boyuan and Wu, Yihan and Wang, Yuyue and Xu, Hongteng and Wang, Yunfeng},
  booktitle={Proceedings of the Computer Vision and Pattern Recognition Conference},
  pages={23369--23378},
  year={2025}
}

@inproceedings{girdhar2023imagebind,
  title={Imagebind: One embedding space to bind them all},
  author={Girdhar, Rohit and El-Nouby, Alaaeldin and Liu, Zhuang and Singh, Mannat and Alwala, Kalyan Vasudev and Joulin, Armand and Misra, Ishan},
  booktitle={Proceedings of the IEEE/CVF conference on computer vision and pattern recognition},
  pages={15180--15190},
  year={2023}
}

@inproceedings{ke2021musiq,
  title={Musiq: Multi-scale image quality transformer},
  author={Ke, Junjie and Wang, Qifei and Wang, Yilin and Milanfar, Peyman and Yang, Feng},
  booktitle={Proceedings of the IEEE/CVF international conference on computer vision},
  pages={5148--5157},
  year={2021}
}

@misc{simeoni2025dinov3,
  title={{DINOv3}},
  author={Sim{\'e}oni, Oriane and Vo, Huy V. and Seitzer, Maximilian and Baldassarre, Federico and Oquab, Maxime and Jose, Cijo and Khalidov, Vasil and Szafraniec, Marc and Yi, Seungeun and Ramamonjisoa, Micha{\"e}l and Massa, Francisco and Haziza, Daniel and Wehrstedt, Luca and Wang, Jianyuan and Darcet, Timoth{\'e}e and Moutakanni, Th{\'e}o and Sentana, Leonel and Roberts, Claire and Vedaldi, Andrea and Tolan, Jamie and Brandt, John and Couprie, Camille and Mairal, Julien and J{\'e}gou, Herv{\'e} and Labatut, Patrick and Bojanowski, Piotr},
  year={2025},
  eprint={2508.10104},
  archivePrefix={arXiv},
  primaryClass={cs.CV},
  url={https://arxiv.org/abs/2508.10104},
}

@inproceedings{huang2024vbench,
  title={Vbench: Comprehensive benchmark suite for video generative models},
  author={Huang, Ziqi and He, Yinan and Yu, Jiashuo and Zhang, Fan and Si, Chenyang and Jiang, Yuming and Zhang, Yuanhan and Wu, Tianxing and Jin, Qingyang and Chanpaisit, Nattapol and others},
  booktitle={Proceedings of the IEEE/CVF Conference on Computer Vision and Pattern Recognition},
  pages={21807--21818},
  year={2024}
}

@inproceedings{teed2020raft,
  title={Raft: Recurrent all-pairs field transforms for optical flow},
  author={Teed, Zachary and Deng, Jia},
  booktitle={European conference on computer vision},
  pages={402--419},
  year={2020},
  organization={Springer}
}

@article{tjandra2025AudioBox-Aesthetics,
  title={Meta audiobox aesthetics: Unified automatic quality assessment for speech, music, and sound},
  author={Tjandra, Andros and Wu, Yi-Chiao and Guo, Baishan and Hoffman, John and Ellis, Brian and Vyas, Apoorv and Shi, Bowen and Chen, Sanyuan and Le, Matt and Zacharov, Nick and others},
  journal={arXiv preprint arXiv:2502.05139},
  year={2025}
}

@inproceedings{radford2023whisper,
  title={Robust speech recognition via large-scale weak supervision},
  author={Radford, Alec and Kim, Jong Wook and Xu, Tao and Brockman, Greg and McLeavey, Christine and Sutskever, Ilya},
  booktitle={International conference on machine learning},
  pages={28492--28518},
  year={2023},
  organization={PMLR}
}

@inproceedings{iashin2024synchformer,
  title={Synchformer: Efficient synchronization from sparse cues},
  author={Iashin, Vladimir and Xie, Weidi and Rahtu, Esa and Zisserman, Andrew},
  booktitle={ICASSP 2024-2024 IEEE International Conference on Acoustics, Speech and Signal Processing (ICASSP)},
  pages={5325--5329},
  year={2024},
  organization={IEEE}
}

@inproceedings{chung2016syncnet,
  title={Out of time: automated lip sync in the wild},
  author={Chung, Joon Son and Zisserman, Andrew},
  booktitle={Asian conference on computer vision},
  pages={251--263},
  year={2016},
  organization={Springer}
}

@misc{Aesthetic,
  author       = {Aesthetic Predictor V2.5},  
  title        = {Aesthetic Predictor V2.5},  
  howpublished = {\url{https://github.com/discus0434/aesthetic-predictor-v2-5}},
   year         = {2024},
}

@article{hu2025ultragen,
  title={UltraGen: High-Resolution Video Generation with Hierarchical Attention},
  author={Hu, Teng and Zhang, Jiangning and Su, Zihan and Yi, Ran},
  journal={arXiv preprint arXiv:2510.18775},
  year={2025}
}

@article{xue2025ultravideo,
  title={UltraVideo: High-Quality UHD Video Dataset with Comprehensive Captions},
  author={Xue, Zhucun and Zhang, Jiangning and Hu, Teng and He, Haoyang and Chen, Yinan and Cai, Yuxuan and Wang, Yabiao and Wang, Chengjie and Liu, Yong and Li, Xiangtai and others},
  journal={arXiv preprint arXiv:2506.13691},
  year={2025}
}

@inproceedings{hu2025high,
  title={High-efficient diffusion model fine-tuning with progressive sparse low-rank adaptation},
  author={Hu, Teng and Zhang, Jiangning and Yi, Ran and Huang, Hongrui and Wang, Yabiao and Ma, Lizhuang},
  booktitle={13th International Conference on Learning Representations, ICLR 2025},
  pages={92066--92078},
  year={2025},
  organization={International Conference on Learning Representations, ICLR}
}

@article{hu2025polyvivid,
  title={PolyVivid: Vivid Multi-Subject Video Generation with Cross-Modal Interaction and Enhancement},
  author={Hu, Teng and Yu, Zhentao and Zhou, Zhengguang and Zhang, Jiangning and Zhou, Yuan and Lu, Qinglin and Yi, Ran},
  journal={arXiv preprint arXiv:2506.07848},
  year={2025}
}

@article{chen2025ivebench,
  title={Ivebench: Modern benchmark suite for instruction-guided video editing assessment},
  author={Chen, Yinan and Zhang, Jiangning and Hu, Teng and Zeng, Yuxiang and Xue, Zhucun and He, Qingdong and Wang, Chengjie and Liu, Yong and Hu, Xiaobin and Yan, Shuicheng},
  journal={arXiv preprint arXiv:2510.11647},
  year={2025}
}

@article{liang2025omniv2v,
  title={OmniV2V: Versatile Video Generation and Editing via Dynamic Content Manipulation},
  author={Liang, Sen and Yu, Zhentao and Zhou, Zhengguang and Hu, Teng and Wang, Hongmei and Chen, Yi and Lin, Qin and Zhou, Yuan and Li, Xin and Lu, Qinglin and others},
  journal={arXiv preprint arXiv:2506.01801},
  year={2025}
}

@inproceedings{wang2024motionctrl,
  title={Motionctrl: A unified and flexible motion controller for video generation},
  author={Wang, Zhouxia and Yuan, Ziyang and Wang, Xintao and Li, Yaowei and Chen, Tianshui and Xia, Menghan and Luo, Ping and Shan, Ying},
  booktitle={ACM SIGGRAPH 2024 Conference Papers},
  pages={1--11},
  year={2024}
}

@inproceedings{hu2025improving,
  title={Improving autoregressive visual generation with cluster-oriented token prediction},
  author={Hu, Teng and Zhang, Jiangning and Yi, Ran and Weng, Jieyu and Wang, Yabiao and Zeng, Xianfang and Xue, Zhucun and Ma, Lizhuang},
  booktitle={Proceedings of the Computer Vision and Pattern Recognition Conference},
  pages={9351--9360},
  year={2025}
}
}

\clearpage

\appendix

\section{Overview}
In this supplementary material, we provide more implementation details, experiment results, including:

\begin{itemize}
    \item Implementation details (Sec.~\ref{sec:implementation details});
    \item Benchmark settings (Sec.~\ref{sec:benchmark_and_metrics});
    \item More quantitative comparisons (Sec.~\ref{sec:more_quantitative_comparisons});
    \item More qualitative comparisons (Sec.~\ref{sec: more qualitative comparisons});
    \item Details about voice clone (Sec.~\ref{sec: details about voice clone});
    \item Audio-driven performance; (Sec.~\ref{sec:audio_driven performance});
    \item More qualitative results (Sec.~\ref{sec: more qualitative results});
    
\end{itemize}
We also provide a \textbf{demo video} and \textbf{project page} in \href{https://sjtuplayer.github.io/projects/Harmony}{\textcolor{magenta}{https://sjtuplayer.github.io/projects/Harmony}}, where the demo video shows the powerful and comprehensive ability of our model in audio-video generation and the project page provides a comparison between the existing methods.

\section{Implementation Details}
\label{sec:implementation details}

\noindent\textbf{Datasets.}
Our training corpus is curated from a diverse range of public and newly collected sources to cover both human speech and environmental sounds.

\textit{1) Human Speech Data:} We aggregate vocal data from multiple open-source datasets, including the TTS-specific Emilia dataset~\cite{he2024emilia}, as well as audio-visual corpora such as OpenHumanVid~\cite{li2025openhumanvid} and SpeakerVid~\cite{zhang2025speakervid}. To ensure high-quality alignment, we employed an audio-visual consistency scoring model to filter this collection, resulting in a high-quality subset of 2 million video clips, each 3-10 seconds in duration. We then utilized the Gemini~\cite{gemini_website} for automated annotation, generating ASR transcripts, descriptive video captions, and captions for any background sounds present in the clips.

\textit{2) Environmental Sound Data:} For environmental sounds, we leverage several established public datasets, including AudioCaps~\cite{kim2019audiocaps} ($\sim$128 hours, manually captioned), Clotho~\cite{drossos2020clotho} ($\sim$31 hours, manually captioned), and WavCaps~\cite{mei2024wavcaps} ($\sim$7,600 hours, automatically captioned). Recognizing the often-suboptimal visual quality of the VGGSound dataset~\cite{chen2020vggsound}, we supplemented our data by collecting an additional 2 million audio-visual clips rich in environmental sounds. These new clips were subsequently annotated using Gemini~\cite{gemini_website} to generate corresponding audio and video captions.

\noindent\textbf{Training Strategy.}
Our training protocol is structured in three distinct stages to ensure stable convergence and high-fidelity generation. For the video branch, we initialize our model with the pre-trained weights of Wan2.2-5B~\cite{wan}. The audio model undergoes a dedicated two-stage pre-training process before the final joint training.

\textit{Stage 1: Foundational Audio Pre-training.}
The audio model is first pre-trained on a balanced 1:1 mixture of our human speech and environmental sound datasets. We train for 100,000 iterations with a global batch size of 1536, using clips with a maximum duration of 10 seconds. During this stage, the reference audio is a randomly selected 1-3 second segment from the ground-truth clip. This phase enables the model to learn to replicate both the timbre and content from the provided reference audio.

\textit{Stage 2: Timbre Disentanglement Finetuning.}
To enable the model to disentangle general acoustic characteristics from specific content, we finetune it using mismatched reference and target content. For human speech, we use cross-utterance data from the same speaker. For environmental sounds, we sample a non-overlapping reference clip from the same long recording as the ground-truth target. This setup compels the model to extract the invariant acoustic signature—be it a speaker's voice or an environmental ambience—from the reference and apply it to the new content dictated by the prompt or transcript. We finetune for an additional 20,000 iterations in this configuration.

\textit{Stage 3: Cross-Task Audio-Visual Training.}
Finally, we proceed to the Cross-Task joint training stage. The full audio-visual model is trained for 10,000 iterations with a batch size of 128, again using a 1:1 mixture of human speech and environmental sound data. Across all training stages, we employ a constant learning rate of 1e-5 for all model parameters.

\noindent\textbf{Hyperparameters.}
During the final cross-task training stage, the balancing weights for our synergistic loss (Eq. ~3) are set to $\lambda_v=0.1$ and $\lambda_a=0.3$. The model is trained using a Flow Matching objective with shift of 5. For inference, we use 40 integration steps with classifier-free guidance (CFG) scales of $s_v=3$ for video and $s_a=2$ for audio. The sampler's shift parameter is also maintained at 5.

\begin{table*}[t]
\centering
\small
\caption{\textbf{Human-speech set comparison} with state-of-the-art joint audio-visual generation models. We evaluate performance across three categories: video quality, audio fidelity, and audio-visual synchronization. Best results are in \textbf{bold}, second-best are \underline{underlined}.}
\label{tab:human_speech compare}
\vspace{-0.1in}
\resizebox{\linewidth}{!}{
    \renewcommand{\arraystretch}{1.2}
    \begin{tabular}{l|ccccc|cccccc|cccc}
    \toprule
    \multirow{2}{*}{Method} & \multicolumn{5}{c|}{Video Quality \& Coherence} & \multicolumn{6}{c|}{Audio Fidelity \& Quality} & \multicolumn{4}{c}{Audio-Visual Synchronization} \\
    \cmidrule(lr){2-6} \cmidrule(lr){7-12} \cmidrule(lr){13-16}
     & AQ$~\uparrow$ & IQ$~\uparrow$ & DD$~\uparrow$ & MS$~\uparrow$ & ID$~\uparrow$ &  PQ$~\uparrow$ & PC$~\downarrow$ & CE$~\uparrow$ & CU$~\uparrow$ & WER$~\downarrow$ & IB-A$~\uparrow$ & Sync-C$~\uparrow$ & Sync-D$~\downarrow$ &DeSync$~\downarrow$&  IB$~\uparrow$ \\
    \midrule

    MM-Diffusion~\cite{ruan2023mmdiffusion} & 0.32 & 0.43 & 0.13 & 0.99  & - & 5.37 & 4.07 & 4.27 & 5.89 & - & - & -  & -&- & 0.12 \\
    
    JavisDiT~\cite{liu2025javisdit} & 0.30 & 0.54 & \textbf{0.28} & 0.99 & 0.35  & 5.34 & 2.16 & 3.61 & 3.92 & 1.00 & 0.14& 1.20 & 12.73 &-& \textbf{0.22} \\

    UniVerse-1~\cite{wang2025universe} & \underline{0.47} & \textbf{0.67} & 0.15 & \underline{0.99} & \underline{0.89}  & 4.28 & 1.86 & 3.84 & 3.91 & 0.23 & \textbf{0.16} & 1.22 & 13.10  &-& 0.16 \\

    Ovi~\cite{low2025ovi} & \textbf{0.48} & \underline{0.65} & 0.17 & 0.99 & 0.88 &  \underline{6.19} & \underline{1.59} & \textbf{5.41} & \textbf{6.21} & \underline{0.19} & 0.10 & \underline{5.13} & \underline{10.38} &- & 0.17 \\
      
    \rowcolor{CornflowerBlue!20}
    \textbf{Harmony (Ours)} & \textbf{0.48} & 0.63 & \underline{0.20} & \textbf{1.00} & \textbf{0.93}  & \textbf{6.20} & \textbf{1.57} & \underline{5.30} & \underline{5.93} & \textbf{0.15} & \underline{0.15} & \textbf{6.51} & \textbf{8.63} & -&\underline{0.18} \\
    \bottomrule
    \end{tabular}
}
\vspace{-0.1in}
\end{table*}

\begin{table*}[t]
\centering
\small
\caption{\textbf{Environment set comparison} with state-of-the-art joint audio-visual generation models. We evaluate performance across three categories: video quality, audio fidelity, and audio-visual synchronization. Best results are in \textbf{bold}, second-best are \underline{underlined}.}
\label{tab:env set compare}
\vspace{-0.1in}
\resizebox{\linewidth}{!}{
    \renewcommand{\arraystretch}{1.2}
    \begin{tabular}{l|ccccc|cccccc|cccc}
    \toprule
    \multirow{2}{*}{Method} & \multicolumn{5}{c|}{Video Quality \& Coherence} & \multicolumn{6}{c|}{Audio Fidelity \& Quality} & \multicolumn{4}{c}{Audio-Visual Synchronization} \\
    \cmidrule(lr){2-6} \cmidrule(lr){7-12} \cmidrule(lr){13-16}
     & AQ$~\uparrow$ & IQ$~\uparrow$ & DD$~\uparrow$ & MS$~\uparrow$ & ID$~\uparrow$ &  PQ$~\uparrow$ & PC$~\downarrow$ & CE$~\uparrow$ & CU$~\uparrow$ & WER$~\downarrow$ & IB-A$~\uparrow$ & Sync-C$~\uparrow$ & Sync-D$~\downarrow$ & DeSync$~\downarrow$ & IB$~\uparrow$ \\
    \midrule

      MM-Diffusion~\cite{ruan2023mmdiffusion} & 0.32 & 0.43 & 0.13 & \underline{0.99}  & - & 5.37 & 4.07 & \textbf{4.27} & 5.89 & - & - & - & - & - & 0.12 \\
JavisDiT~\cite{liu2025javisdit} & 0.37 & 0.55 & 0.33 & \underline{0.99} & 0.45  & 5.64 & \textbf{2.29} & 3.06 & 5.14 & - & \underline{0.18} & - & - & \underline{0.94} & 0.16 \\
UniVerse-1~\cite{wang2025universe} & 0.57 & \textbf{0.68} & 0.16 & \textbf{1.00} & \underline{0.92}  & 6.14 & \underline{2.30} & 3.20 & 5.46 & - & 0.04 & - & - & 1.10 & 0.07 \\
Ovi~\cite{low2025ovi} & \underline{0.62} & \underline{0.66} & \underline{0.44} & \underline{0.99} & \textbf{0.93}  & \underline{6.45} & 2.46 & 3.78 & \underline{5.98} & - & \textbf{0.20} & - & - & 1.06 & \underline{0.20} \\

\rowcolor{CornflowerBlue!20}
\textbf{Harmony (Ours)} & \textbf{0.64} & 0.65 & \textbf{0.56} & 0.98 & 0.90  & \textbf{6.53} & 2.68 & \underline{4.12} & \textbf{6.22} & - & 0.14 & - & - & \textbf{0.70} & \textbf{0.21} \\
    \bottomrule
    \end{tabular}
}
\end{table*}

\begin{table*}[t]
\centering
\small
\caption{\textbf{Complex set comparison} with state-of-the-art joint audio-visual generation models. We evaluate performance across three categories: video quality, audio fidelity, and audio-visual synchronization. Best results are in \textbf{bold}, second-best are \underline{underlined}.}
\label{tab:complex sec compare}
\vspace{-0.1in}
\resizebox{\linewidth}{!}{
    \renewcommand{\arraystretch}{1.2}
    \begin{tabular}{l|ccccc|cccccc|cccc}
    \toprule
    \multirow{2}{*}{Method} & \multicolumn{5}{c|}{Video Quality \& Coherence} & \multicolumn{6}{c|}{Audio Fidelity \& Quality} & \multicolumn{4}{c}{Audio-Visual Synchronization} \\
    \cmidrule(lr){2-6} \cmidrule(lr){7-12} \cmidrule(lr){13-16}
     & AQ$~\uparrow$ & IQ$~\uparrow$ & DD$~\uparrow$ & MS$~\uparrow$ & ID$~\uparrow$  & PQ$~\uparrow$ & PC$~\downarrow$ & CE$~\uparrow$ & CU$~\uparrow$ & WER$~\downarrow$ & IB-A$~\uparrow$ & Sync-C$~\uparrow$ & Sync-D$~\downarrow$ & DeSync$~\downarrow$ & IB$~\uparrow$ \\
    \midrule
      MM-Diffusion~\cite{ruan2023mmdiffusion} & 0.32 & 0.43 & 0.13 & \underline{0.99}  & - & 5.37 & 4.07 & \underline{4.27} & \textbf{5.89} & - & - & - & - & - & 0.12 \\
      JavisDit~\cite{liu2025javisdit} & 0.34 & 0.50 & \textbf{0.54} & 0.98 & 0.33  & 5.40 & 2.26 & 2.91 & 4.56 & 1.00 & \textbf{0.09} & 0.58 & 10.50 & 1.32 & \underline{0.17} \\
      UniVerse-1~\cite{wang2025universe} & \underline{0.52} & \underline{0.65} & \underline{0.42} & \underline{0.99} & 0.85  & \underline{6.14} & \underline{2.23} & 3.85 & 5.15 & \underline{0.25} & 0.00 & 0.72 & \underline{8.32} & \textbf{1.09} & 0.14 \\
    Ovi~\cite{low2025ovi} & 0.60 & 0.63 & 0.41 & \underline{0.99} & \underline{0.88}  & 5.94 & 2.33 & 4.14 & \underline{5.33} & 0.79 & \underline{0.06} & \underline{2.94} & 8.86 & 1.21 & \textbf{0.18} \\
    \rowcolor{CornflowerBlue!20}
      \textbf{Harmony (Ours)} & \textbf{0.64} & \textbf{0.66} & 0.32 & \textbf{1.00} & \textbf{0.91}  & \textbf{6.43} & \textbf{1.90} & \textbf{4.76} & 4.86 & \textbf{0.15} & \underline{0.06} & \textbf{4.70} & \textbf{6.43} & \underline{1.13} & \textbf{0.18} \\


    \bottomrule
    \end{tabular}
}
\end{table*}

\begin{table}[t]
\centering
\small
\caption{\textbf{Chinese speech comparison} with state-of-the-art models, focusing on audio fidelity (WER) and audio-visual synchronization. Best results are in \textbf{bold}, second-best are \underline{underlined}.}
\label{tab:chinese_set_comparison}
\vspace{-0.1in}
\resizebox{0.95\linewidth}{!}{ 
    \renewcommand{\arraystretch}{1.2}
    \begin{tabular}{l|c|ccc}
    \toprule
     Method& WER$~\downarrow$ & Sync-C$~\uparrow$ & Sync-D$~\downarrow$ & IB$~\uparrow$ \\
    \midrule
    JavisDiT~\cite{liu2025javisdit} & 4.84 & 1.27 & 12.63 & \underline{0.20} \\
    UniVerse-1~\cite{wang2025universe} & \underline{2.32} & 0.91 & 11.02 & \textbf{0.22} \\
    Ovi~\cite{low2025ovi} & 9.10 & \underline{4.45} & \underline{10.79} & \underline{0.20} \\
      \rowcolor{CornflowerBlue!20}
      \textbf{Harmony (Ours)} & \textbf{0.92} & \textbf{5.05} & \textbf{9.38} & \textbf{0.22} \\

    \bottomrule
    \end{tabular}
}
\end{table}

\begin{figure*}[t]
    \centering
    \includegraphics[width=1.0\textwidth]{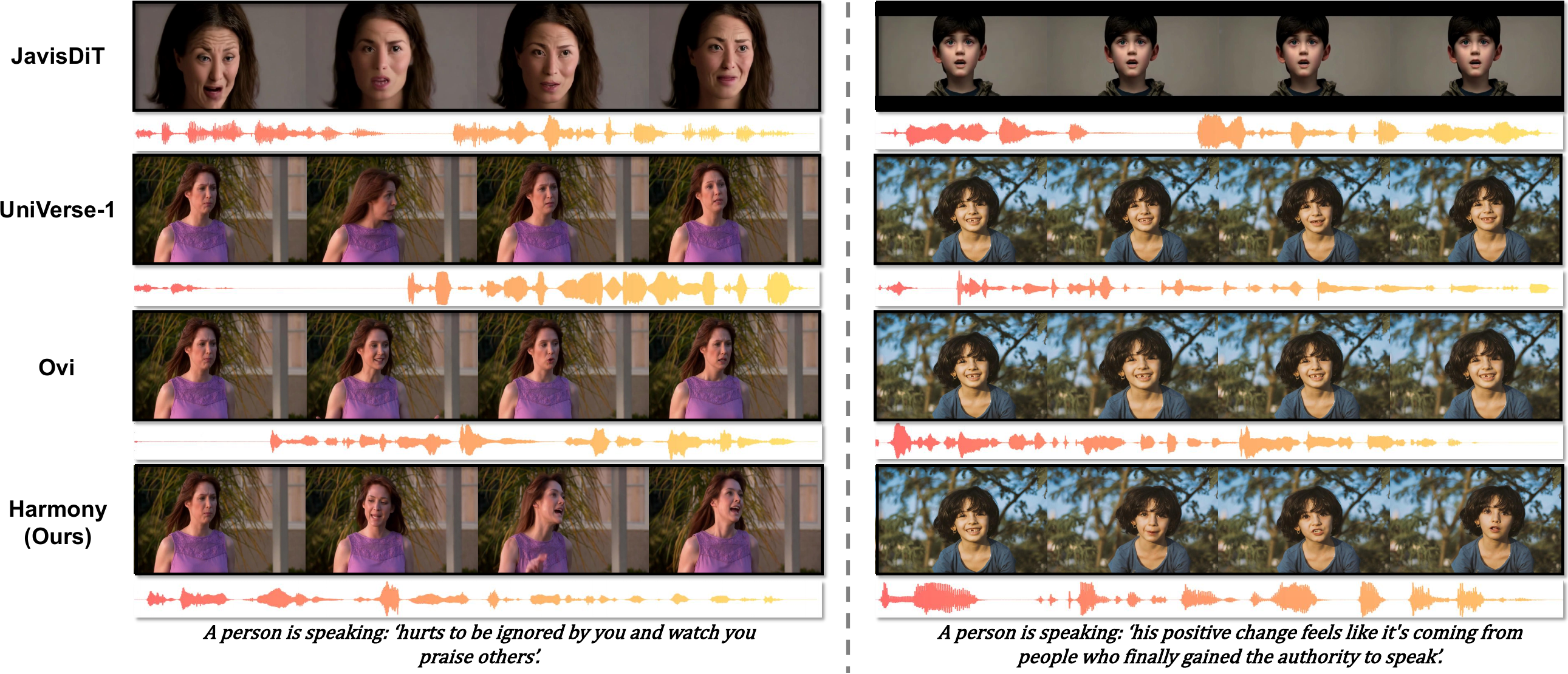}
    \vspace{-0.1in}
    \caption{More comparison on human-speech video generation.}
    \vspace{-0.1in}
    \label{fig:more compare human}
\end{figure*}

\begin{figure*}[t]
    \centering
    \includegraphics[width=1.0\textwidth]{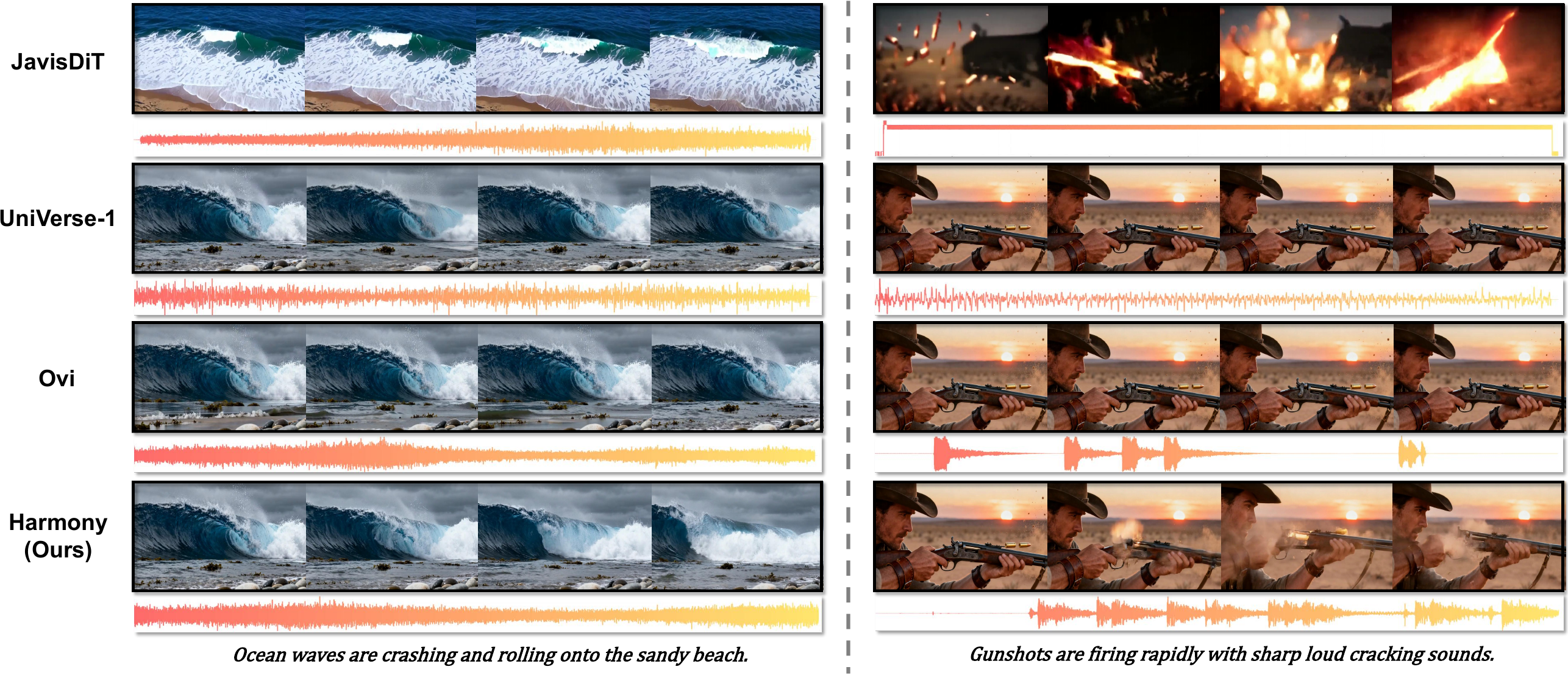}
    \vspace{-0.1in}
    \caption{More comparison on environment-sound video generation.}
    \vspace{-0.1in}
    \label{fig:more compare env}
\end{figure*}

\section{Benchmark Settings}
\label{sec:benchmark_and_metrics}

\subsection{The Harmony-Bench Dataset}

Existing benchmarks for audio-visual generation are inadequate for comprehensive evaluation. JavisBench~\cite{liu2025javisdit} lacks evaluation for human speech, while Verse-Bench~\cite{wang2025universe} is hampered by low-quality labels and a limited focus on audio-visual synchronization. To enable a more rigorous and holistic assessment, we construct and introduce \textbf{Harmony-Bench}. This new benchmark features 150 meticulously designed test cases, organized into three progressively challenging subsets (50 items each). It is specifically crafted to disentangle and systematically evaluate a model's semantic consistency and temporal synchronization across diverse and complex acoustic scenarios.


\begin{itemize}
    \item \textbf{Ambient Sound-Video Generation.} This subset is designed to assess the model's ability to generate non-speech acoustic events that are precisely synchronized with corresponding visual dynamics. The 50 test cases feature synthetically constructed scenarios, enabling the creation of complex audio-visual interactions that are difficult to capture or isolate in real-world recordings. The model is conditioned on a detailed \texttt{audio\_caption} and a separate \texttt{video\_caption}. Evaluation centers on audio fidelity, temporal synchrony, and the semantic consistency between the generated audio and visual events.

    \item \textbf{Speech-Video Generation.} This 50-item subset assesses the fidelity of speech synthesis and lip synchronization. To test for robustness and multilingual generalization, it includes a balanced mix of 25 real-world and 25 AI-synthesized samples, driven by transcripts in both English (\texttt{spoken\_word\_en}) and Chinese (\texttt{spoken\_word\_zh}). The \texttt{video\_caption} is deliberately kept minimal (e.g., "a man is speaking"), compelling the model to derive lip movements and facial expressions directly from the transcript's content. Key evaluation criteria are speech intelligibility, naturalness, and the precision of lip-audio synchronization.

    \item \textbf{Complex Scene: Ambient + Speech.} Representing the most challenging scenario, this subset evaluates the model's capacity to simultaneously generate and synchronize both speech and ambient sounds within a unified, complex scene. Each of the 50 test cases is constructed to feature co-occurring audio-visual events, requiring the model to process a combination of inputs: a transcript (\texttt{spoken\_word\_en}), an ambient sound description (\texttt{audio\_caption}), and a visual scene description (\texttt{video\_caption}). The evaluation critically examines the model's ability for sound source separation and mixing (e.g., maintaining speech clarity over a background door-closing sound). Furthermore, it assesses multi-modal temporal alignment: speech must synchronize with lip movements, while ambient sounds must align with their corresponding visual actions.
\end{itemize}

To provide a comprehensive evaluation on this benchmark, we adopt a suite of automated metrics designed to assess three key aspects: 1) Visual Quality and Coherence, 2) Audio Fidelity, and 3) Audio-Visual Synchronization and Consistency.

\subsection{Evaluation Metrics}
\label{sec:evaluation_metrics}

To comprehensively assess model performance on Harmony-Bench, we employ a suite of automated metrics targeting three core aspects of audio-visual quality.

\noindent\textbf{Visual Quality and Coherence.} We evaluate the visual quality and temporal consistency of the generated videos using the following metrics:
\begin{itemize}
    \item \textbf{Aesthetic and Imaging Quality.} We assess \textbf{aesthetic quality (AQ)} and \textbf{imaging quality (IQ)} using the pre-trained aesthetic-predictor-v2-5\cite{Aesthetic} and MUSIQ\cite{ke2021musiq} models, respectively.
    \item \textbf{Motion Dynamics.} Temporal coherence is evaluated through \textbf{Dynamic Degree (DD)} and \textbf{Motion Smoothness (MS)}\cite{huang2024vbench}. We employ RAFT\cite{teed2020raft} to quantify the magnitude of motion and a pre-trained video frame interpolation model to evaluate motion smoothness.
    \item \textbf{Identity Consistency (ID).} For subject-specific generation, we measure ID by computing the mean DINOv3\cite{simeoni2025dinov3} feature similarity between a reference image and all generated frames.
\end{itemize}

\noindent\textbf{Audio Fidelity and Quality.} The quality of the generated audio is measured by:
\begin{itemize}
    \item \textbf{AudioBox-Aesthetics.}\cite{tjandra2025AudioBox-Aesthetics} We employ this model to evaluate perceptual quality across four dimensions: \textbf{Production Quality (PQ)}, \textbf{Production Complexity (PC)}, \textbf{Content Enjoyment (CE)}, and \textbf{Content Usefulness (CU)}.
    \item \textbf{Word Error Rate (WER).} For speech synthesis, accuracy is measured by WER. We transcribe the generated audio using Whisper-large-v3\cite{radford2023whisper} and compare it against the ground-truth transcript.
    \item \textbf{IB-A Score.} Semantic alignment between the generated audio and the text prompt is quantified using the IB-A Score\cite{girdhar2023imagebind}.
\end{itemize}

\noindent\textbf{Audio-Visual Synchronization.} The critical capability of joint generation is assessed through synchronization metrics:
\begin{itemize}
    \item \textbf{Sync-C \& Sync-D.} Lip-sync accuracy is explicitly measured using these two established metrics\cite{chung2016syncnet}.
    \item \textbf{DeSync Score.} Predicted by Synchformer\cite{iashin2024synchformer}, this score quantifies the temporal misalignment (in seconds) between the audio and video streams.
    \item \textbf{ImageBind (IB) Score.} Following~\cite{girdhar2023imagebind}, we use the IB score to assess overall audio-visual consistency by computing the cosine similarity between their respective feature embeddings.
\end{itemize}

\section{More Quantitative Comparisons}
\label{sec:more_quantitative_comparisons}

In this section, we present detailed quantitative comparisons against state-of-the-art methods for joint audio-video generation, including Ovi~\cite{low2025ovi}, UniVerse-1~\cite{wang2025universe}, JavisDiT~\cite{liu2025javisdit}, and MM-Diffusion~\cite{ruan2023mmdiffusion}. Our evaluation spans multiple challenging test sets, with results for environmental sounds and complex audio scenes presented in Tables~\ref{tab:human_speech compare}--\ref{tab:complex sec compare}. Across these diverse datasets, our model consistently demonstrates superior performance. A key observation is our model's superior video dynamism compared to competitors. For instance, while UniVerse-1 and Ovi sometimes achieves a favorable Identity Distance (ID) score, this is often a consequence of generating static or nearly static videos, where frame-to-frame identity is trivially high but fails to capture the scene's intended motion. Crucially, our method consistently achieves the lowest Word Error Rate (WER) and the best scores on audio-visual synchronization metrics. This combination of high fidelity, strong dynamism, and precise alignment underscores our model's robustness in generating coherent and realistic content for complex scenes.

Furthermore, we specifically assess the cross-lingual capabilities of the models on a dedicated Chinese speech test set, with key results summarized in Table~\ref{tab:chinese_set_comparison}. The results highlight a significant performance gap. Our model achieves a substantially lower WER and markedly better synchronization scores. It is worth noting that the standard WER metric is not perfectly optimized for the tokenization of the Chinese language; therefore, the relative performance between models serves as the most meaningful indicator. The pronounced improvement in both WER and synchronization metrics strongly validates the effectiveness and superiority of our approach for cross-lingual audio-visual speech generation.

\section{More Qualitative Comparisons}
\label{sec: more qualitative comparisons}

In this section, we present further qualitative comparisons of our method against state-of-the-art approaches: Ovi~\cite{low2025ovi}, UniVerse-1~\cite{wang2025universe}, and JavisDiT~\cite{liu2025javisdit}. We focus on two challenging scenarios: synchronized human speech and dynamic environmental sounds. We exclude MM-Diffusion~\cite{ruan2023mmdiffusion} from this analysis as it is designed for unconditional generation and is therefore not directly comparable.

\noindent\textbf{Comparisons on Human Speech.}
As illustrated in Figure~\ref{fig:more compare human}, our model demonstrates superior performance in audio-visual speech generation. Competing methods like Ovi and UniVerse-1 tend to produce static or minimally dynamic video frames, resulting in a "talking head" effect with little natural movement. In contrast, our model generates high-fidelity video with fluid, naturalistic motion. The accompanying audio is clear and, most importantly, precisely synchronized with the lip movements, resulting in a significantly more coherent and believable output.

\noindent\textbf{Comparisons on Environmental Sounds.}
We further evaluate performance on generating dynamic environmental sounds in Figure~\ref{fig:more compare env}, where the shortcomings of other methods are even more pronounced. JavisDiT struggles in this domain, producing low-quality video and unstable audio; for instance, in the "gunfire" example, its generated audio waveform is highly irregular and fails to represent the acoustic event convincingly. UniVerse-1 and Ovi frequently generate static or partially static scenes. A clear example is the "ocean waves" case, where the main waves remain frozen while only the water surface shows minimal movement. This lack of dynamism is compounded by poor audio-visual synchronization, where the sound of crashing waves does not align with the visual content. In stark contrast, our method excels in all aspects: it generates high-quality, dynamic videos with realistic motion, and the synthesized audio is both high-fidelity and precisely synchronized with the visual events, delivering a cohesive and immersive audio-visual experience.

\begin{figure*}[t]
    \centering
    \includegraphics[width=0.9\textwidth]{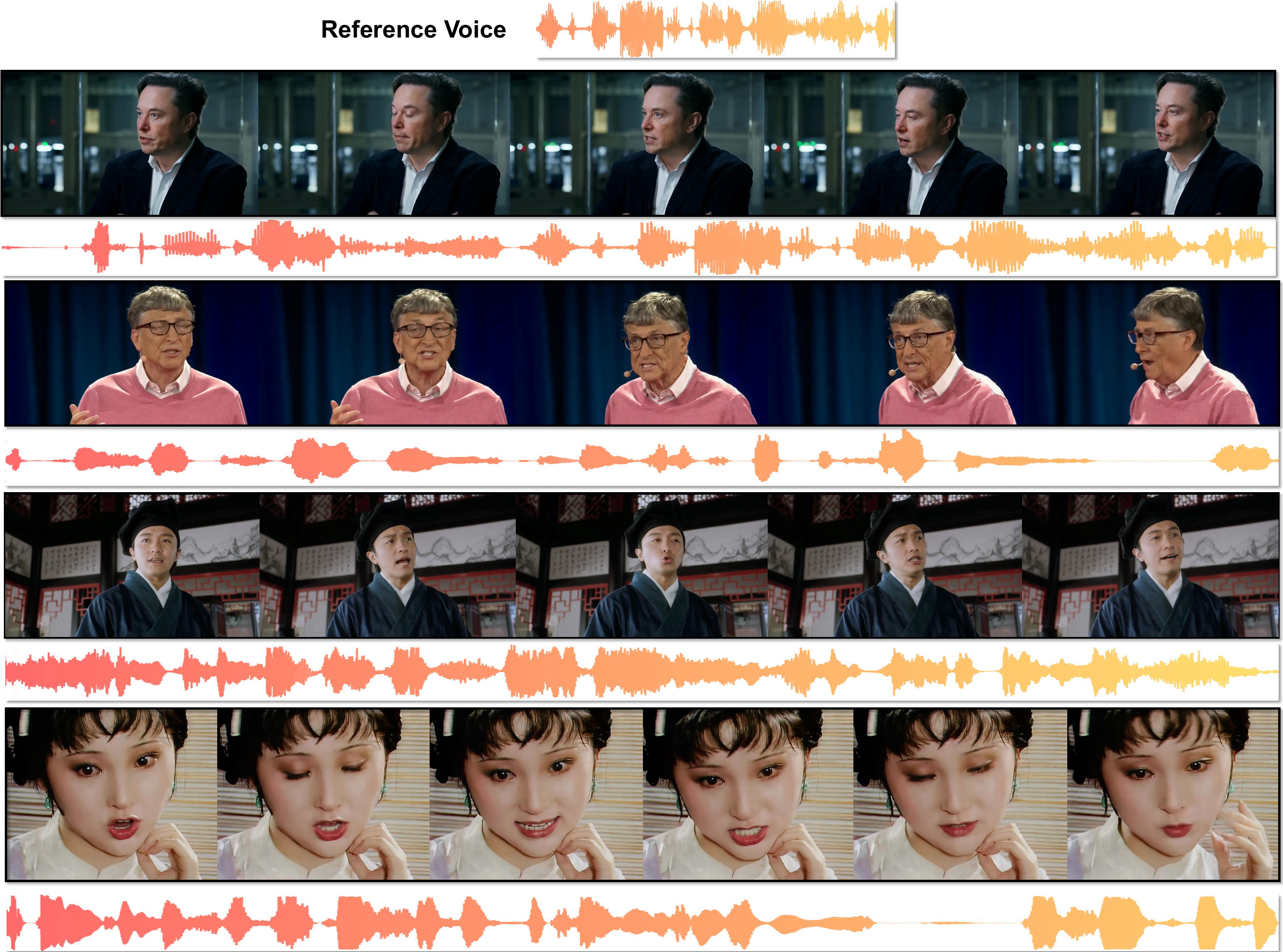}
    \caption{Visualization of the voice-clone results of our model.}
    \label{fig:voice clone}
\end{figure*}

\section{Details about Voice Clone}
\label{sec: details about voice clone}

In this section, we provide additional details on the voice cloning capability of our model, which is achieved through the use of a reference audio input, $A_r$. The mechanism begins by processing a short reference audio clip (typically 1-3 seconds) containing the desired voice timbre with our pre-trained audio VAE encoder~\cite{cheng2025mmaudio}. This yields a compact latent representation, $z_r$, which effectively captures the unique, time-invariant characteristics of the speaker's voice while discarding the original phonetic content. As described in our main methodology, this reference latent $z_r$ is then prepended to the noisy target audio latent $z_{a,t}$ during each step of the denoising process. By conditioning the MM-DiT on this fixed reference latent, the model is guided to synthesize new speech---based on the phonetic content from the transcript $T_s$---in the desired target voice.

To qualitatively validate the effectiveness of this approach, we provide examples in Figure~\ref{fig:voice clone}. The figure demonstrates that our model can successfully clone a variety of distinct voice timbres onto newly generated speech content. Importantly, this high-fidelity voice cloning is achieved without degrading the visual quality of the generated video. The lip movements remain precisely synchronized with the cloned audio, and the overall facial expressions and video coherence are maintained at a high level. This highlights the model's ability to disentangle audio timbre from other generation aspects, enabling robust voice cloning within a coherent audio-visual output.

\begin{figure*}[t]
    \centering
    \includegraphics[width=0.9\textwidth]{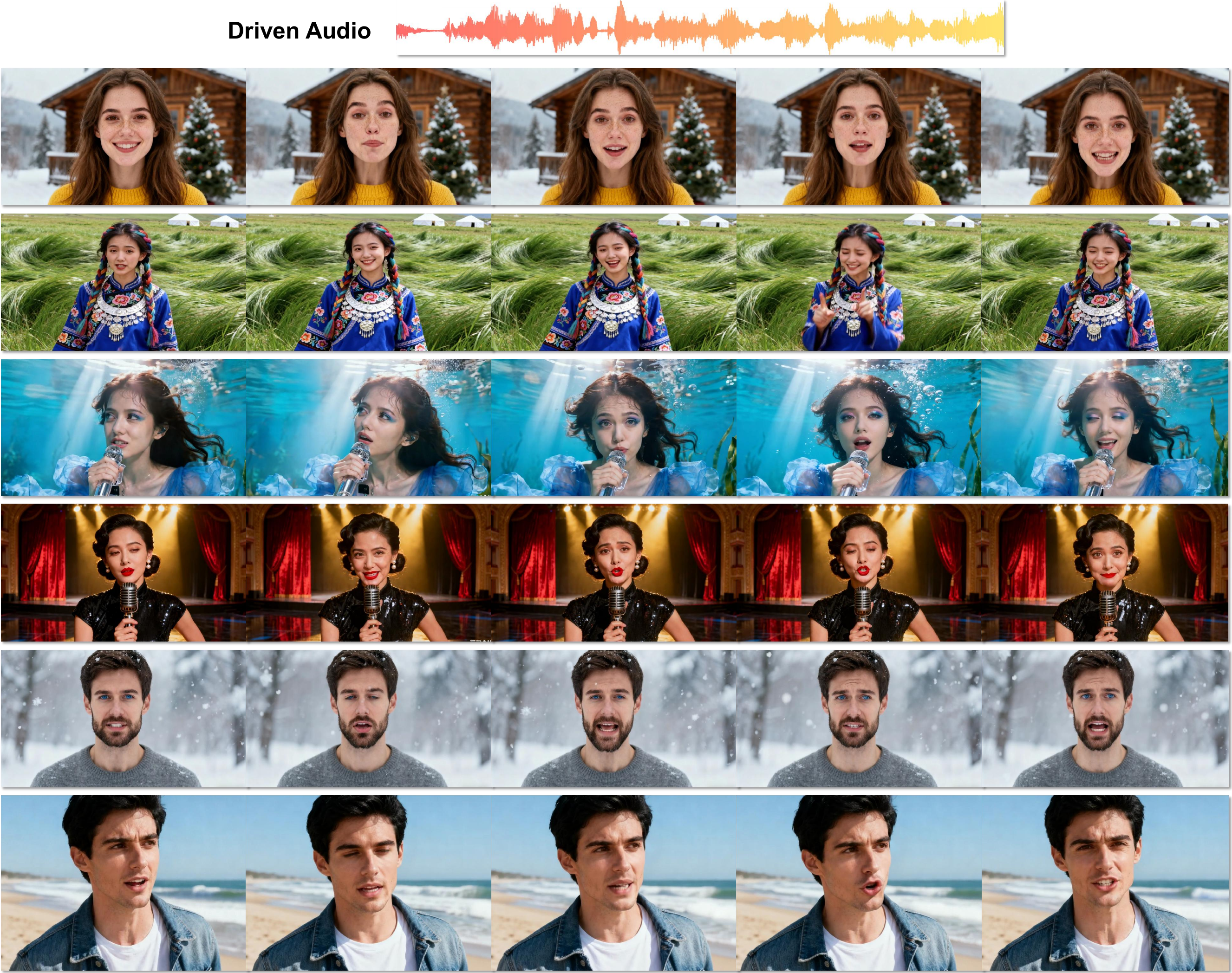}
    \caption{Visualization of the audio-driven results of our model.}
    \label{fig:audio_driven performance}
\end{figure*}

\section{Audio-Driven Performance}
\label{sec:audio_driven performance}

As detailed in our main paper, our Cross-Task Synergy Training strategy is fundamental to the model's performance. A key component of this strategy is the inclusion of a deterministic, audio-driven video generation task, represented by the loss term $\mathcal{L}_{\text{driven}}^\text{audio}$. During training, this task explicitly requires the video branch to generate video conditioned on the clean, non-noisy audio latent $z_{a,0}$ (i.e., the audio latent at timestep $t_a=0$). By directly optimizing for this objective, our model is inherently equipped with the ability to perform high-fidelity audio-driven video synthesis at inference time, making it a native capability rather than an emergent one.

To demonstrate the effectiveness of this native capability, we present qualitative results for audio-driven video generation in Figure~\ref{fig:audio_driven performance}. The figure showcases examples where video is generated solely from a target speech audio clip. The results exhibit high visual quality, characterized by natural facial expressions and coherent head movements. More importantly, the lip movements are precisely and accurately synchronized with the nuances of the input speech, validating the strong audio-visual alignment instilled by our training approach. This confirms that our Cross-Task Synergy strategy not only enhances joint generation but also directly enables high-fidelity, single-modality-driven applications.

\section{More qualitative results}
\label{sec: more qualitative results}

To further demonstrate the capabilities and robustness of our model, we present additional qualitative results organized into three key areas: generating high-quality human speech videos, rendering diverse artistic styles, and synthesizing complex ambient sounds.

\begin{figure*}[t]
    \centering
    \includegraphics[width=0.98\textwidth]{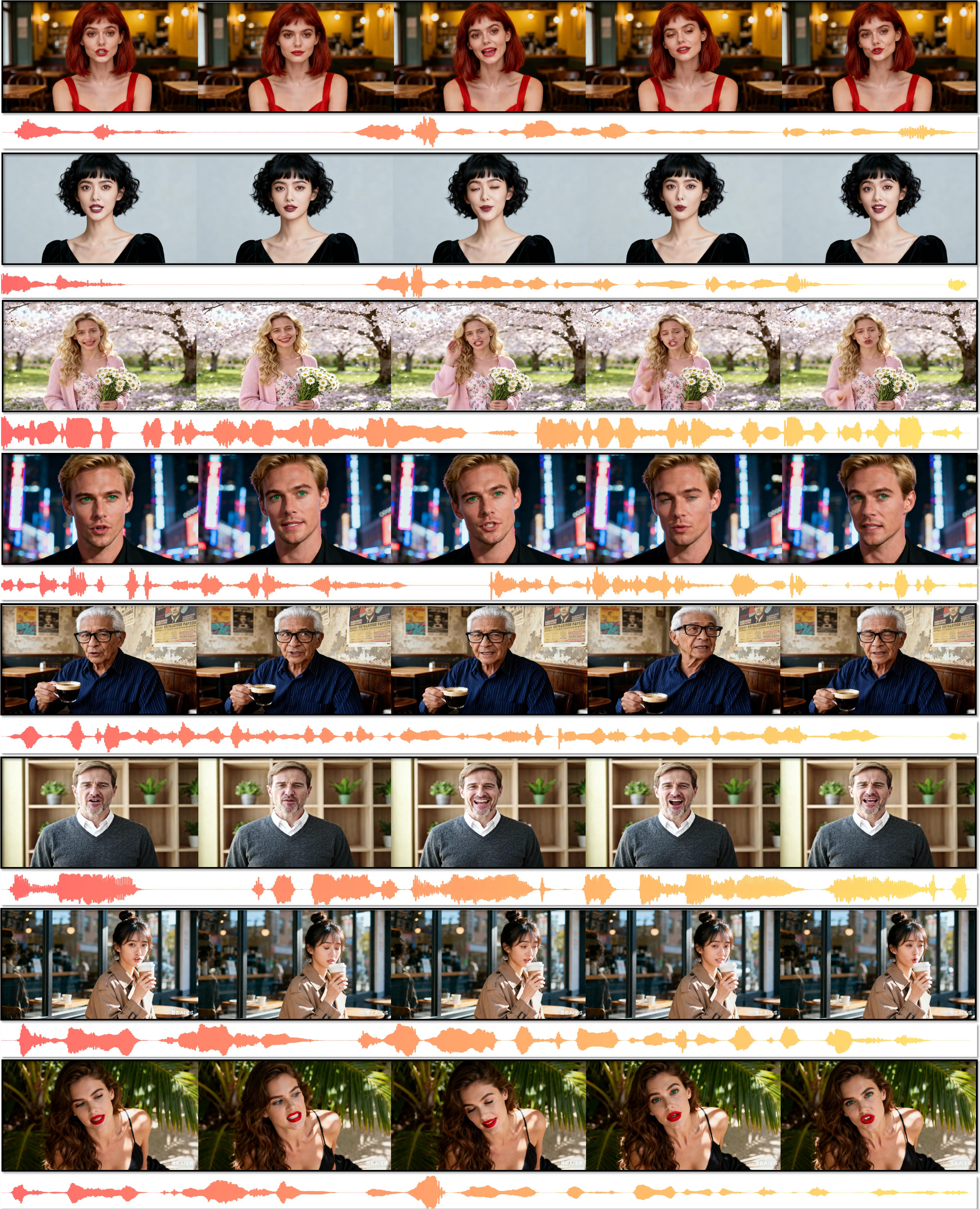}
    \caption{More results on human-speech video generation.}
    \label{fig:more_results_on_speech}
\end{figure*}

\noindent\textbf{More results on human speech.} 
First, we showcase additional results on generating human speech videos in Figure~\ref{fig:more_results_on_speech}. These examples highlight the model's ability to produce highly realistic talking heads with natural facial expressions and coherent movements. The synthesized speech is characterized by its clarity and natural prosody, capturing a range of vocal tones. Crucially, we maintain precise lip synchronization across all examples, which is fundamental for creating believable human speech. These results reinforce our model's core capability in generating high-quality, well-synchronized audio-visual speech content across various identities.

\begin{figure*}[t]
    \centering
    \includegraphics[width=0.98\textwidth]{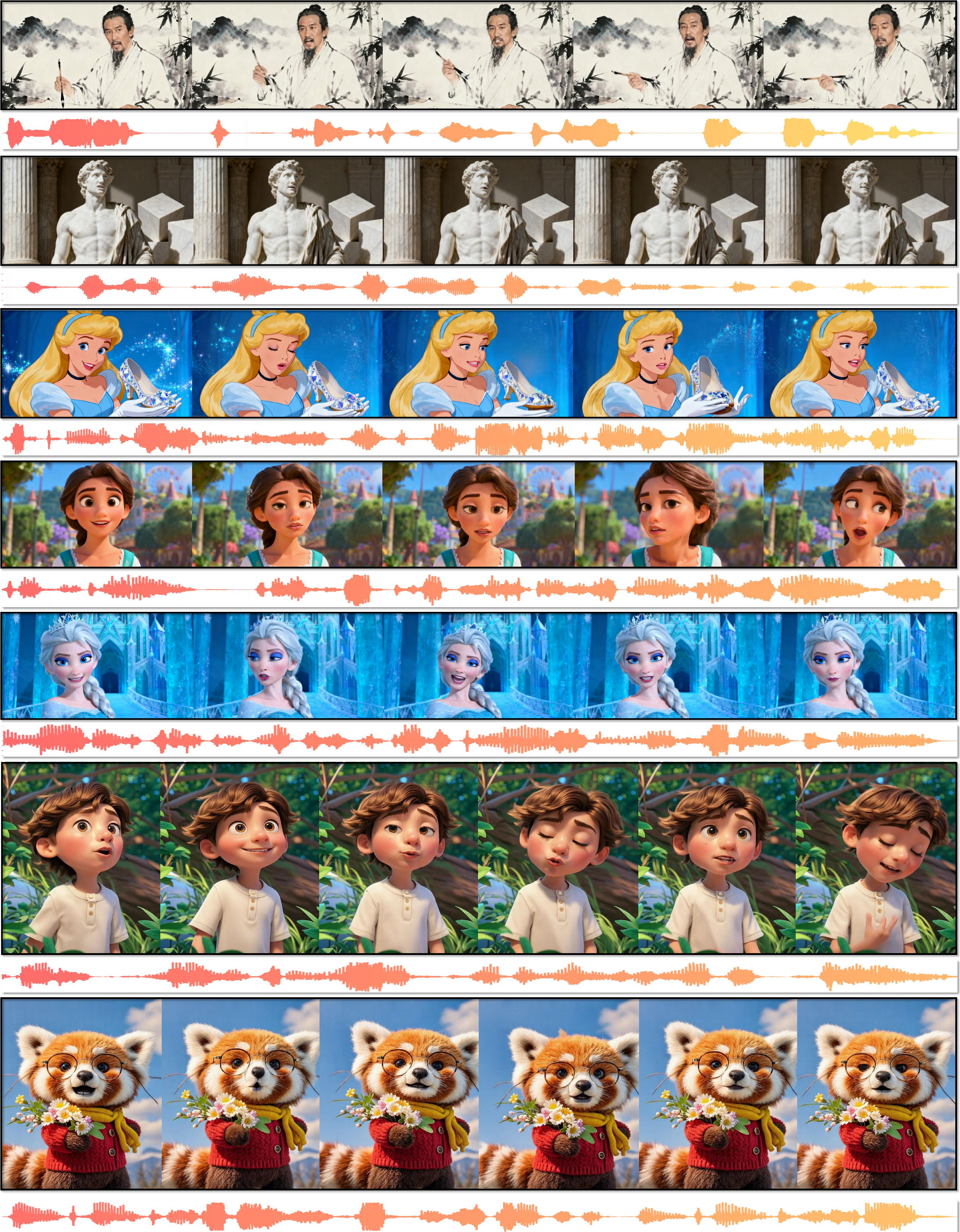}
    \caption{Visualization of speech-video generation in diverse style.}
    \label{fig:diverse stlyle performance}
\end{figure*}

\noindent\textbf{Diverse visual styles.} 
Beyond photorealism, a key strength of our model is its capacity to generate video content across a wide spectrum of artistic styles. As illustrated in Figure~\ref{fig:diverse stlyle performance}, our model can produce outputs in distinct aesthetics such as Disney-style animation and traditional ink wash painting. These stylized generations maintain high visual quality, characterized by sharp details, vibrant colors, and temporally coherent motion consistent with the target aesthetic. This demonstrates the model's flexibility in capturing and rendering complex artistic attributes.

\begin{figure*}[t]
    \centering
    \includegraphics[width=0.98\textwidth]{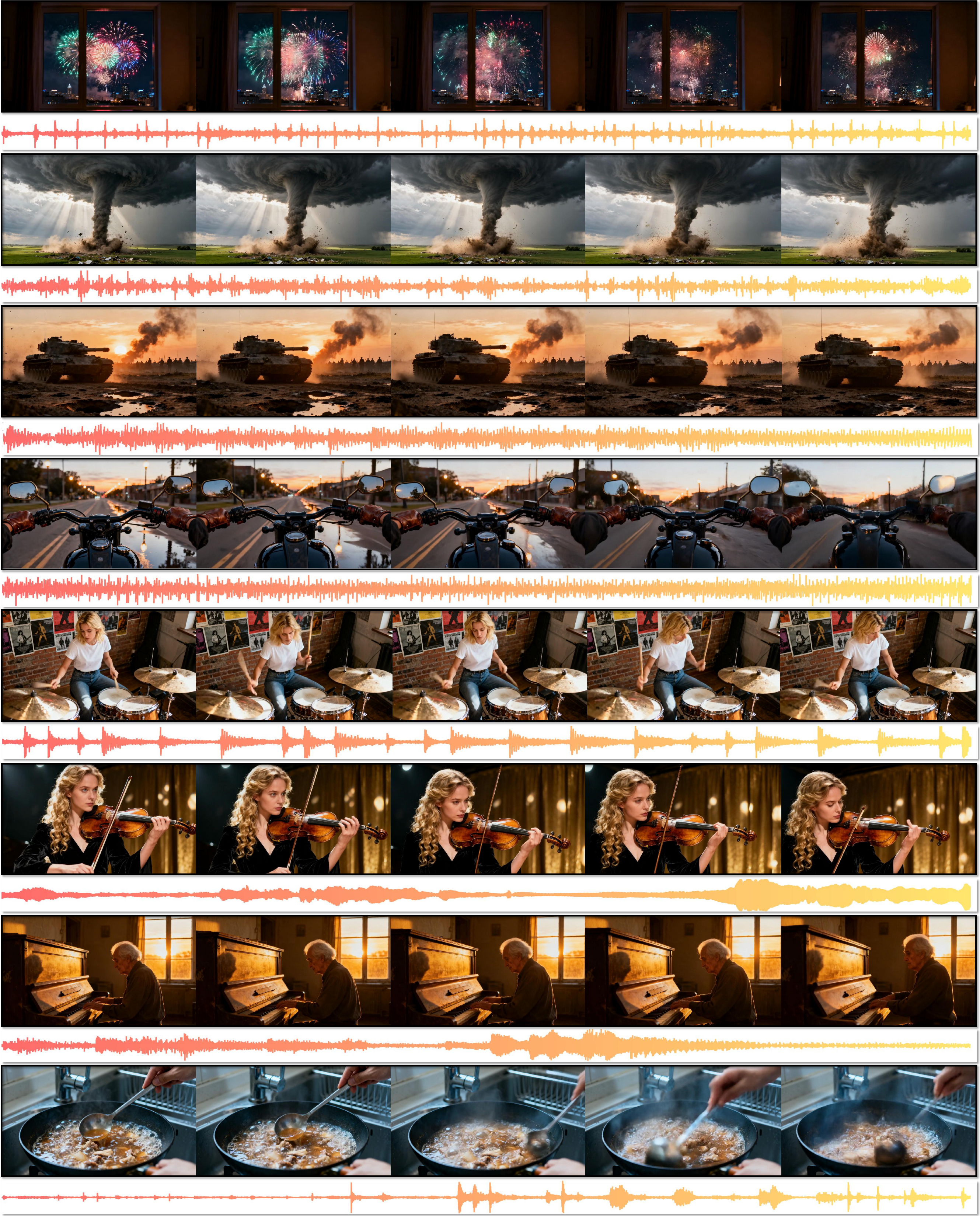}
    \caption{More results on ambient-sound video generation.}
    \label{fig:more_results_on_env}
\end{figure*}

\noindent\textbf{Diverse Ambient Sounds.} 
Our model demonstrates a remarkable capability to generate a wide spectrum of ambient sounds, extending beyond simple environmental noise. As illustrated in Figure~\ref{fig:more_results_on_env}, it can produce diverse and complex acoustic events—from the sharp, percussive bursts of fireworks to the structured harmonies of music. Crucially, each sound is rendered with high fidelity and meticulously synchronized with its corresponding visual source. This ability to construct rich, thematically consistent auditory environments validates our model's strength in enhancing the overall visual narrative.

Collectively, these examples validate our model's comprehensive generation capabilities. From producing highly synchronized human speech to rendering diverse artistic styles and creating rich, context-aware ambient soundscapes, our model demonstrates remarkable versatility. The ability to master these distinct yet complementary domains underscores its potential for creating highly expressive and immersive audio-visual content, pushing the boundaries beyond conventional generation methods.


\clearpage

\end{document}